\documentclass[runningheads]{llncs}

\usepackage{eccv}

\usepackage{color}
\usepackage{xcolor}
\usepackage{graphicx}
\usepackage{subcaption}
\usepackage{pifont}
\usepackage{multirow}
\usepackage{xspace}
\usepackage{caption}
\usepackage{amsmath}
\usepackage{enumitem}
\usepackage{wrapfig,lipsum}
\usepackage{diagbox}
\usepackage{booktabs}
\usepackage{textcomp}
\usepackage{gensymb}

\def\ourmethod{{\textit{ColorPeel}}\xspace}
\newcommand{\minisection}[1]{\vspace{0.005in} \noindent {\bf #1}}

\newcommand{\model}{\epsilon_\theta}

\newcommand{\conditioner}{\tau_\theta}
\newcommand{\expec}{\mathbb{E}}
\newcommand{\encoder}{\mathcal{E}}
\newcommand{\decoder}{\mathcal{D}}

\newcommand{\textprompt}{\mathcal{P}}
\newcommand{\textembedding}{\mathcal{C}}

\newcommand{\tokenset}{\mathcal{V}}
\newcommand{\concept}{\mathcal{V}^{*}}

\newcommand{\colortoken}{\mathit{c}^{*}}
\newcommand{\colortokeni}{\mathit{c}^{*}_i}

\newcommand{\shapetoken}{\mathit{s}^{*}}
\newcommand{\shapetokeni}{\mathit{s}^{*}_i}
\newcommand{\shapetokenj}{\mathit{s}^{*}_j}
\newcommand{\texturetoken}{\mathit{t}^{*}}
\newcommand{\materialtoken}{\mathit{m}^{*}}

\newcommand{\crossattn}{\mathcal{A}}

\newcommand{\mlp}{\mathit{l}}

\newcommand{\inputimage}{\mathcal{I}}
\newcommand{\quotes}[1]{``#1''}

\usepackage{eccvabbrv}

\usepackage{graphicx}
\usepackage{booktabs}

\usepackage[accsupp]{axessibility}  

\usepackage{hyperref}

\usepackage{orcidlink}

\begin{document}
\title{ColorPeel: Color Prompt Learning with Diffusion Models via  Color and Shape Disentanglement} 

\titlerunning{ColorPeel}

\author{Muhammad Atif Butt\inst{1,2}\orcidlink{0000-0002-4799-6488} \and
Kai Wang\inst{1}\thanks{Corresponding Author}\orcidlink{0000-0002-9605-8279}\and
Javier Vazquez-Corral\inst{1,2}\orcidlink{0000-0003-0414-7096} \and
Joost van de Weijer\inst{1,2}\orcidlink{0000-0002-9656-9706}}

\authorrunning{M.A.~Butt et al.}

\institute{Computer Vision Center, Spain \and
Universitat Autonoma de Barcelona \\
\email{\{mabutt,kwang,jvazquez,joost\}@cvc.uab.es}
}

\maketitle

\begin{figure}[h]
  \centering
  \includegraphics[width=0.999\textwidth]{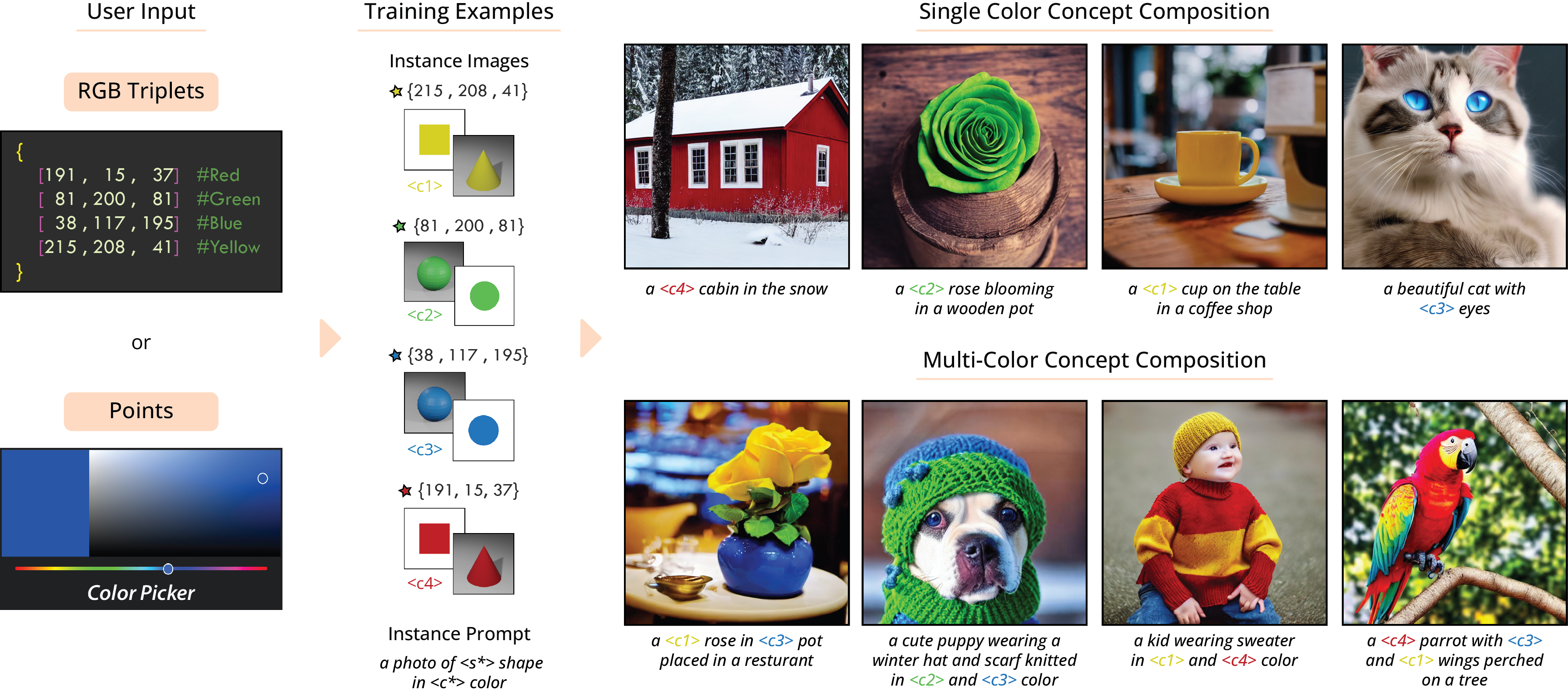}
  \caption{
  Overview of our \ourmethod for personalized color prompt learning. Given the RGB triplets or color coordinates, \ourmethod generates basic 2D or 3D geometries with target colors for color learning. This facilitates the disentanglement of color and shape concepts, allowing for precise color usage in image generation.
  }
  \label{fig:teaser}
  \vspace{-6mm}
\end{figure}
\begin{abstract}
Text-to-Image (T2I) generation has made significant advancements with the advent of diffusion models. These models exhibit remarkable abilities to produce images based on textual prompts. Current T2I models allow users to specify object colors using linguistic color names. However, these labels encompass broad color ranges, making it difficult to achieve precise color matching. To tackle this challenging task, named \textit{color prompt learning}, we propose to learn specific color prompts tailored to user-selected colors. Existing T2I personalization methods tend to result in color-shape entanglement. To overcome this, we generate several basic geometric objects in the target color, allowing for color and shape disentanglement during the color prompt learning. Our method, denoted as \ourmethod, successfully assists the T2I models to \textit{peel off} the novel color prompts from these colored shapes. In the experiments, we demonstrate the efficacy of \ourmethod in achieving precise color generation with T2I models. Furthermore, we generalize \ourmethod to effectively learn abstract attribute concepts, including textures, materials, etc. Our findings represent a significant step towards improving precision and versatility of T2I models, offering new opportunities for creative applications and design tasks. Our project is available at \href{https://moatifbutt.github.io/colorpeel/}{https://moatifbutt.github.io/colorpeel/}.
\end{abstract}

\section{Introduction}
\label{sec:intro}

Text-to-Image (T2I) generation has seen enormous improvements since the arrival of diffusion models~\cite{chang2023muse,deepfloyd,ramesh2022dalle2,ramesh2021zero,saharia2022imagen}. These models, which are trained on an enormous amount of pairs of images and captions, have remarkable ability to generate images guided by user text prompts. In combination with inversion methods~\cite{mokady2022null,han2023ProxNPI,miyake2023NPI,tang2023iterinv,tang2024locinv}, these models can be used to edit real-world images~\cite{hertz2022prompt,tumanyan2022plug,brooks2022instructpix2pix,parmar2023zero,kai2023DPL}, e.g., by replacing objects, modifying attribute intensity, changing background, etc. In this paper, we focus on the capabilities of diffusion models to generate objects of a precise color. This capability plays a pivotal role in design, fashion and art, where it is important to generate objects in the exact color envisaged by the user~\cite{singh2006impact}. 

Current T2I diffusion models~\cite{Rombach_2022_CVPR_stablediffusion,ho2022imagen,podell2023sdxl} allow users to specify color of generated object using color names~\cite{berlin1991basic, van2009learning}, which are linguistic labels like ‘red’, ‘green’, and ‘blue’. However, these color names encompass a wide range of object reflectance, and even when using more precise color names like ‘beige’ or ‘light green’, the generated results may not precisely match the intended color. This discrepancy arises as language represents color in a \textit{discrete} manner, whereas color is a \textit{continuous} concept. Therefore, opting for an approach that enables users to select an exact color from a color palette is more desirable. 
This approach will provide users with precise control over colors of the generated objects.

\begin{figure}[t]
  \centering
  \includegraphics[width=0.9555\textwidth]{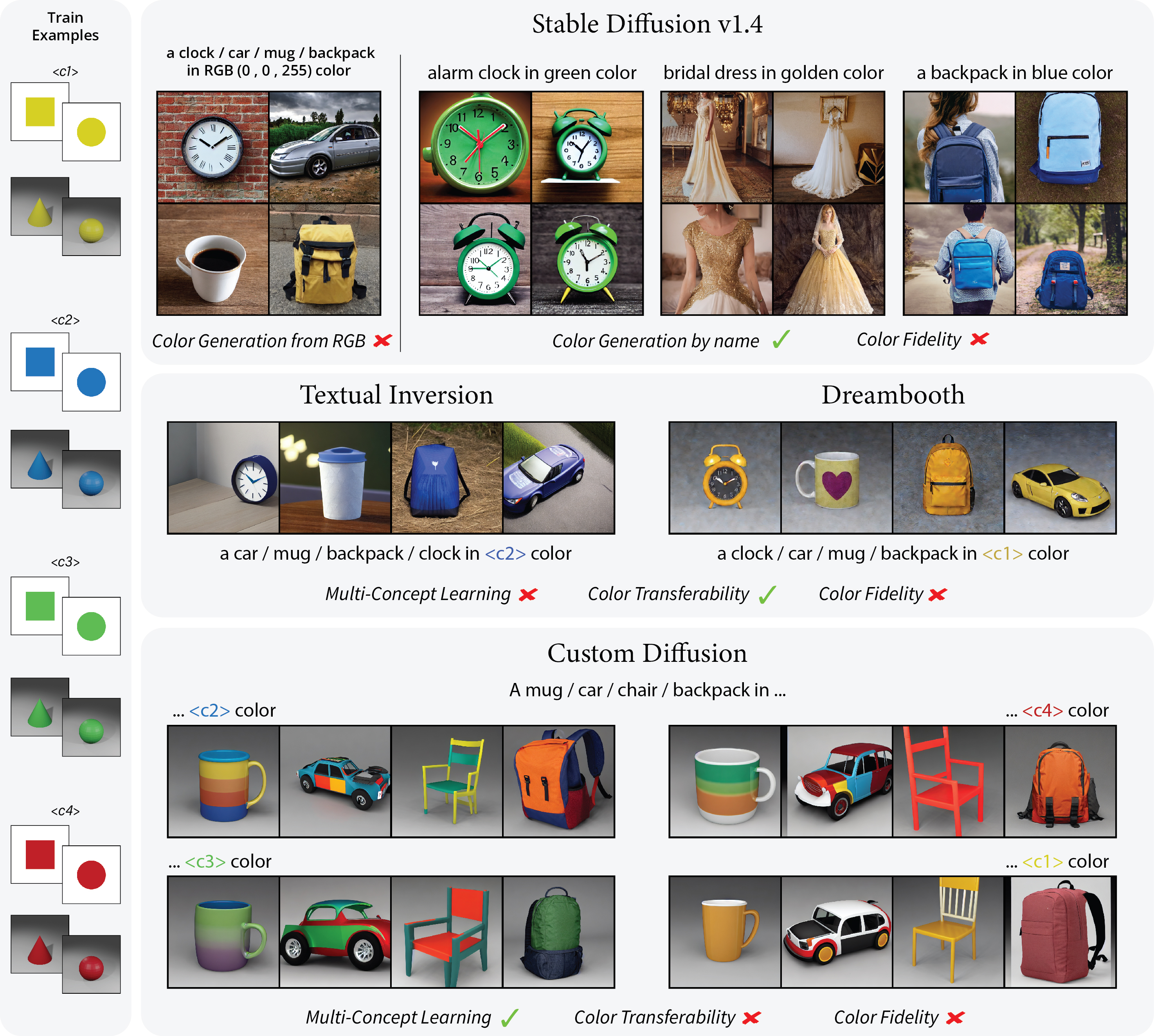}
\vspace{-1mm}
  \caption{Analyzing Color Fidelity and Transferability. Given RGB values (of blue color) in the text prompt, \textbf{Stable Diffusion} fails to generate desired objects in specified colors and also lacks consistency in color fidelity when provided with specific color names. Comparatively, seminal new concept learning methods \textbf{Textual Inversion} and \textbf{Dreambooth} generate text-guided objects in specified colors; however, these are single concept learning baselines and also fail to generate consistent colored objects. \textbf{Custom Diffusion} --- multi-concept learning baseline, inter-mixes the colors while also reducing the sample variation, which leads to unintended outcomes.}
\vspace{-4mm}
  \label{fig:baselines}
\end{figure}

To address the challenge of precise color generation, we set out to learn specific \textit{color prompts} for the color selected by a user. These colors can then be used to generate objects of the same color. As a solution to the \textit{color prompt learning} task, current T2I personalization methods~\cite{ruiz2022dreambooth,textual_inversion,kumari2022customdiffusion,hiper2023} offer a naive transfer learning approach, by which we can learn color prompt from a patch entirely in the target color.
However, we demonstrate that these baselines fail to produce satisfactory results because they do not correctly disentangle color from shape ({as evident in Fig.~\ref{fig:baselines}}). 
Moreover, attempting to input the exact RGB values into T2I models result in unsatisfactory results, as demonstrated in Rich-Text~\cite{ge2023richtext} 
(see Fig.~\ref{fig:baselines_lc} in the supplementary).

To tackle this issue, we propose to generate a set of basic geometric objects with the target color (in 2D or 3D shapes as shown in Fig.~\ref{fig:teaser}) and then use these instance images to learn the color prompts. Furthermore, we apply a new \textit{cross-attention alignment} loss that further improves disentanglement. Subsequently, we obtain a series of tokens representing the target colors and shapes. This disentanglement-based learning approach, termed \ourmethod, effectively assists the T2I diffusion model in acquiring the ability of \textit{color prompt learning} by \textit{peeling off} the color attributes from geometric shape objects. 

In the experiments, we first demonstrate that the learned color prompts can effectively generate objects in the desired target colors, whether these colors are coarse-grained or fine-grained. We then evaluate the precision of the generated colors by computing color difference metrics and conducting user studies, which confirm that \ourmethod outperforms various baseline methods. Additionally, we illustrate how these prompts can be utilized for image editing by recoloring objects from input images. 
We also explore interpolation between various learned color prompts.
Finally, we investigate the generalizability of \ourmethod by extending the training scheme to learn textures and materials from user input. To summarize, we have the following contributions:
\begin{itemize}
    \item This paper is the first to tackle the \textit{color prompt learning} problem, a crucial aspect in content creation. This addresses the need of T2I model users to generate precise colors, which is vital in various creative endeavors.
    \item We introduce \ourmethod, an effective solution accompanied by a novel \textit{cross-attention alignment} loss. 
    This method is designed to tackle the challenges of color learning by disentangling colors and shapes from the automatically generated geometric objects with target colors.
    
    \item Our method outperforms other T2I approaches by a large margin on quantitative results and a user study. We further show that our method can be extended to textures and material properties.
\end{itemize}

\section{Related Work}
\label{sec:background}

\minisection{Transfer learning for T2I models.} 
Transfer learning for T2I models is also referred to \textit{T2I model adaptation} or \textit{personalized generation}. 
It aims at adapting a given model to a \textit{new concept} by given images from the users and bind the new concept with a unique token. As a result, the adaptation model can generate various renditions for the new concept guided by text prompts. Depending on whether the adaptation method is fine-tuning the T2I model, they are categorized into two main streams: 
(1) \textit{Fine-tuning the T2I model.} One of the most representative methods is DreamBooth~\cite{ruiz2022dreambooth}, where the pretrained T2I model such that it learns to bind a unique identifier with that specific subject given 3$\sim$5 images. 
Custom Diffusion (CD)~\cite{kumari2022customdiffusion} and other approaches~\cite{gal2023designing,han2023svdiff,zhang2022sine,chen2023suti,shi2023instantbooth,Cones2023} are also following this pipeline and improving the generation quality.
(2) \textit{Freeze the T2I model.} Another stream focuses on learning new concept tokens instead of fine-tuning the generative models.
Textual Inversion (TI)~\cite{textual_inversion} is a pioneering work focusing on finding new pseudo-words by conducting personalization in the text embedding space.
Following works~\cite{dong2022dreamartist,daras2022multiresolution,voynov2023ETI,hiper2023} continue to improve this technique stream.

Despite existing T2I model adaptation methods have been successful in learning new concepts from a set of relevant images, they have overlooked the user's requirement to generate objects with custom-defined colors, and have thus struggled to meet this challenge. In this paper, our objective is to develop a learning scheme for these methods, equipping them with the ability for \textit{color prompt learning}. This enhancement expands the potential of existing T2I adaptation methods in artistic creation. 
While there have been several papers addressing the extraction of multiple concepts from a single image~\cite{avrahami2023breakascene,vinker2023concept_decomposition,lopes2023material,motamed2023lego,yeh2024texturedreamer}, these efforts predominantly concentrate on extracting concrete concepts implicitly. For example, Break-a-Scene~\cite{avrahami2023breakascene} aligns the cross-attention maps with segmentation masks to learn new concepts separately for each object. Concept Decomposition~\cite{vinker2023concept_decomposition} disentangles one object implicitly into several concepts.
However, as they cannot ensure clean disentanglement between the concrete and abstract concepts, they are not directly applicable to the \textit{color prompt learning}.

\minisection{Text-Guided Image Editing.} With the recent progress of T2I models~\cite{chang2023muse,gafni2022make,hong2022sag,ramesh2022dalle2,ramesh2021zero,saharia2022imagen}, various text-guided image methods~\cite{hertz2023delta_DDS,meng2022sdedit,mokady2022null,zhang2023forgedit,chen2023fec,li2023stylediffusion} are explored to adopt such T2I models for controllable image editing. Imagic~\cite{kawar2022imagic} and P2P~\cite{hertz2022prompt} attempt structure-preserving editing via Stable Diffusion (SD) models.  InstructPix2Pix~\cite{brooks2022instructpix2pix} is an extension of P2P by allowing human-like instructions for image editing.
To make P2P capable of handling real images, Null-Text inversion (NTI)~\cite{mokady2022null} proposed to update \textit{null-text} embeddings for accurate reconstruction to accommodate with classifier-free guidance~\cite{ho2022classifier}. Following approaches~\cite{parmar2023zero,tumanyan2022plug,couairon2023diffedit} deal with text-guided image editing through various techniques, they can be further explored in survey papers~\cite{huang2024diffusion_image_edit_survey,direct_inversion_2023,basu2023editval}. Nonetheless, existing methods heavily depend on the expressive power of the underlying T2I diffusion models and struggle to efficiently control the color attributes of generated objects for tasks like image editing or generation. In this paper, in addition to learning specific tokens for user-requested novel colors, we also conduct experiments to generate  images using newly learned color tokens.

\section{Method}
\label{sec:method}
\begin{figure}[t]
  \centering
  \includegraphics[width=0.999\linewidth]{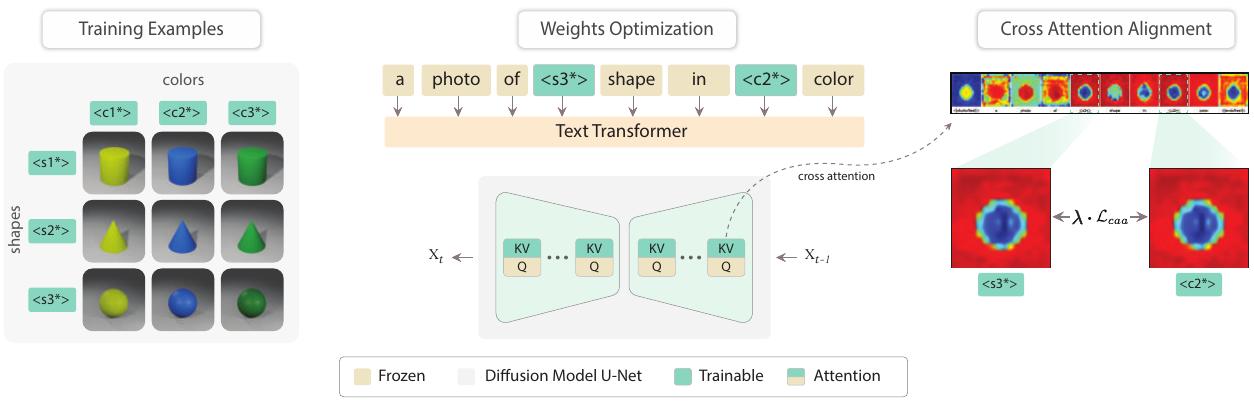}
  \vspace{-6mm}
  \caption{Illustration of our method \ourmethod. Firstly, instance images along with the templates are generated, given the user-provided RGB or color coordinates. Next, we introduce new modifier tokens, i.e., $\shapetokeni$ and $\colortokeni$ which correspond to shapes and colors to ensure the disentanglement of shape from color. Following Custom Diffusion, the key and value projection matrices in the diffusion model cross-attention layers are optimized along with the modifier tokens while training. To improve learning, we introduce \textit{cross attention alignment} to enforce the color and shape cross-attentions.
  }
  \vspace{-4mm}
  \label{fig:method}
\end{figure}

In this section, we describe our method \ourmethod to achieve  \textit{color prompt learning}. An illustration of \ourmethod is shown in Fig.~\ref{fig:method}.

\subsection{Preliminaries}
\noindent{\bf Latent Diffusion Models.}
In this paper, we use Stable Diffusion v1.4 as the backbone model, which is built on the Latent Diffusion Model (LDM)~\cite{Rombach_2022_CVPR_stablediffusion}. The model is composed of two main components: an autoencoder and a diffusion model. The encoder $\encoder$ from the autoencoder component of the LDMs maps an image $\inputimage$ into a latent code $z_0=\encoder(\inputimage)$ and the decoder reverses the latent code back to the original image as $\decoder(\encoder(\inputimage)) \approx \inputimage$.
The diffusion model can be conditioned on class labels, segmentation masks or textual input. Let $\tau_\theta(y)$ be the conditioning mechanism which maps a condition $y$ into a conditional vector for LDMs. The LDM model is updated by the noise reconstruction loss:

\begin{equation}
L_{LDM} = \expec_{z_0 \sim \encoder(x), y, \epsilon \sim \mathcal{N}(0, 1)} \underbrace{\left( \Vert \epsilon - \model(z_{t},t, \conditioner(y)) \Vert_{2}^{2} \right)}_{\mathcal{L}_{rec}}.
\label{eq:loss}
\end{equation}

The neural backbone $\model$ is a conditional UNet~\cite{ronneberger2015unet} which predicts the added noise. In particular, text-guided diffusion models aim to generate an image from a random noise $z_T$ and a conditional input prompt $\textprompt$. To distinguish from general conditional notation in LDMs, we itemize textual condition as $\textembedding=\tau_\theta(\textprompt)$. 

The cross-attention maps in the Stable Diffusion UNet module (SD-UNet) between textual input and images, can be obtained from $\model(z_t,t,\textembedding)$.
They are computed from deep features of the noisy image $f_{z_t}$ which are projected to a query matrix $Q_t=\mlp_Q (f_{z_t})$, and textual embedding which is projected to a key matrix $K = \mlp_K (\textembedding)$. Then the attention map is computed according to:
\begin{equation}
\mathcal{A}_t=softmax(Q_t \cdot K^T / \sqrt{d})
\end{equation}
where $d$ is the latent dimension, and the cell $[\mathcal{A}_t]_{ij}$ defines the weight of the $j$-th token on the $i$-th token.

\noindent{\bf T2I model transfer learning.}
Given a pretrained T2I diffusion model, T2I adaptation methods~\cite{kumari2022customdiffusion,ruiz2022dreambooth,textual_inversion} aim to embed a new concept in the model given few images along with text description. The fine-tuned model should retain its prior knowledge, allowing novel generations with new concept based on the text prompt. As a common practice, novel token learning via text encoding is applied.

To personalize the target concept images, a corresponding text caption is necessary. In scenarios where the target concept represents a unique instance within a broader category, a new modifier token $\concept$ is introduced.
During training, $\concept$ is initialized with a rare occurring token embedding and optimized with customized losses.
Furthermore, for fine-tuning based transfer learning methods, they also update conditional SD-UNet partially (like CD~\cite{kumari2022customdiffusion}) or fully (like DB~\cite{ruiz2022dreambooth}) to obtain better learning performance.

\subsection{Color Prompt Learning}
\label{subsec:cpl}
Despite the wide application of existing adaptation methods in learning new concepts, they mainly focus on prompt learning for concrete concepts and generally ignore changing attributes, like color attributes. In this paper, we refer to this task as \textit{color prompt learning}. We observe that the naive approaches cannot solve this task ({as shown in Fig.~\ref{fig:baselines}}). These methods fail to disentangle the color information from the training images. 

Therefore, we propose to generate a series of geometric shapes with target colors to disentangle (or \textit{peel off}) the target colors from the shapes. By jointly learning on multiple color-shape images, we found that the method can successfully disentangle the color and shape concepts. For simplicity, we further denote the target color concepts as $\colortoken$ and shape concepts as $\shapetoken$.
There should be at least two shapes $\shapetokeni,\shapetokenj$ with the same target color $\colortoken$ for the model to analogize the color attribute. 
In this paper, we consider two sets of shapes 
(see Fig.~\ref{fig:dataset_coarse} in supplementary) one set of 2D shapes and another of 3D shapes. Since the 3D shapes undergo physical transformation such as shading and shadow effects, which are also present in the generated images, we expect these to yield improved color prompts compared to those  learned from 2D shapes. 
The images are corresponding to prompts $\textprompt$ like: “A photo of $\shapetokeni$ filled with $\colortoken$”, “A photo of $\shapetokenj$ shape with $\colortoken$ color”, etc.
More information on the prompt templates are provided is Section~\ref{subsec:prompt_templates}.

\begin{figure}[t]
  \centering
  \includegraphics[width=0.999\textwidth]{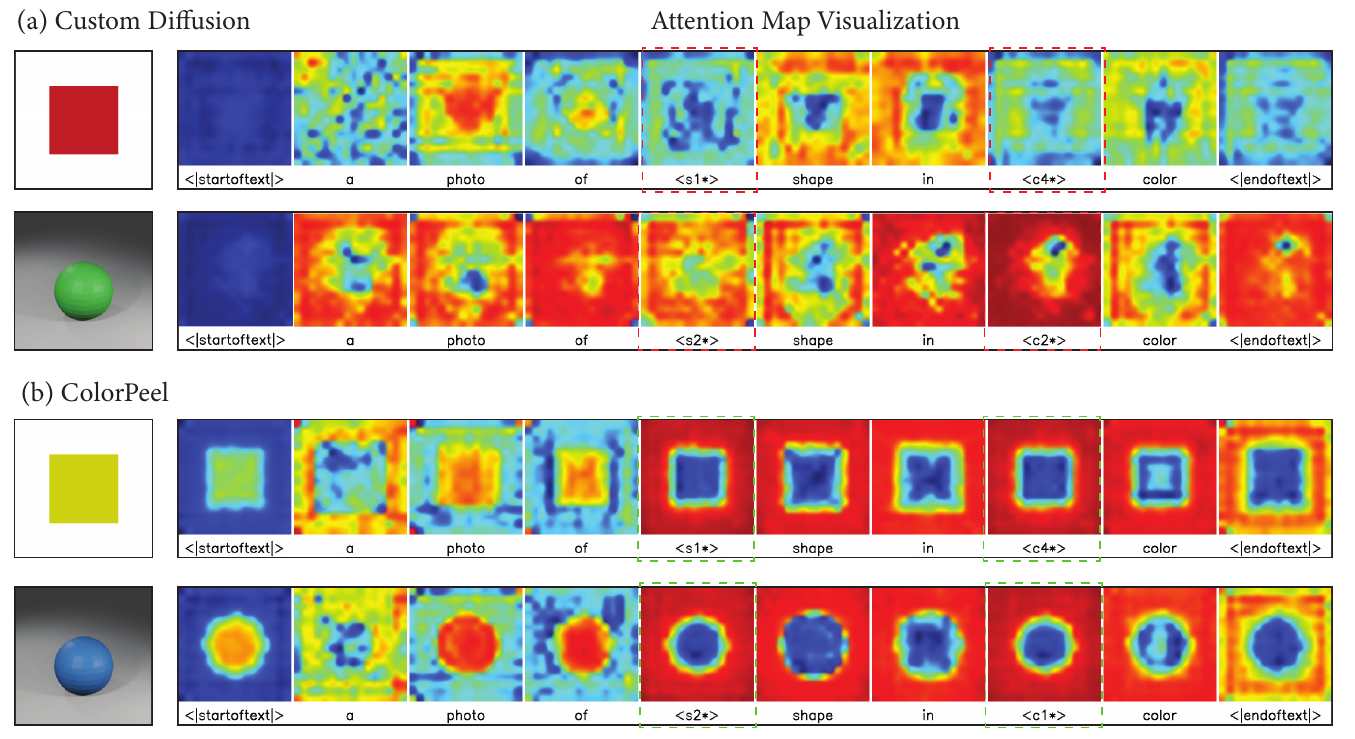}
  \vspace{-6mm}
  \caption{Cross attention visualization. We compare the cross attention maps from the last timestep of Custom Diffusion and \ourmethod. Our method precisely learns color from the given concept while distinctively avoiding the overlapping with background, which is one of the main reasons for color inter-mixing in the baseline.
  }
  \vspace{-5mm}
  \label{fig:cross_attn}
\end{figure}

In order to learn the novel color token embeddings $\tokenset^{\colortoken}$, we randomly sample an image from our small training set, which depict our target color in various shapes. We directly optimize new tokens ($\tokenset^{\colortoken}$, $\tokenset^{\shapetoken}$) and the SD-UNet (optional for Custom Diffusion, DreamBooth, etc.) by minimizing the LDM loss as defined in Eq.~\ref{eq:loss}. In this way, our optimization goal can be defined as
\begin{equation}
\concept = \underset{\tokenset}{\arg\min} \   \expec_{z_0 \sim \encoder(x), y, \epsilon \sim \mathcal{N}(0, 1)} \; \mathcal{L}_{rec} 
\label{eq:token_loss}
\end{equation}
This is optimized by re-using the same training strategy as the original LDM model. As such, we aim to encourage the learned embedding to capture fine visual details unique to the concept.

\noindent{\bf Cross-Attention Alignment (CAA).} Using only Eq.~\ref{eq:token_loss} results in some improvement over the baseline, but the generated colors do not accurately depict the target color and sometimes struggle to correctly disentangle color from shape. 
By visualizing the cross-attention maps from the SD-UNet modules ({as shown in Fig.~\ref{fig:cross_attn}} and further illustrated in Fig.~\ref{fig:supp_attn_maps} in the supplementary), we hypothesize the misalignment between the color and shape attentions are the root of this unsatisfactory performance. Intuitively, we propose the \textit{cross-attention alignment} (CAA) loss to achieve agreement between these cross-attentions, as defined by the cosine similarity between the cross-attention maps:
\begin{align}
\mathcal{L}_{caa} = 1-\mathrm{cos} (\crossattn_t^{\colortoken},\crossattn_t^{\shapetoken})
\label{eq:caa_loss}.
\end{align}
This loss is motivated by DPL~\cite{kai2023DPL}, however, DPL is reversed to minimize overlap of attention between different objects. 
Our final optimization function is:
\begin{equation}
\concept = \underset{\tokenset}{\arg\min} \   \expec_{z_0 \sim \encoder(x), y, \epsilon \sim \mathcal{N}(0, 1)} \; \Big[ \mathcal{L}_{rec} + \lambda \cdot \mathcal{L}_{caa} \Big]
\label{eq:final_loss}
\end{equation}
where $\lambda$ is the trade-off hyperparameter. 
In conclusion, in this paper, \quotes{disentanglement} refers to the process of \textit{decoupling} shapes and colors from a set of auto-generated colored geometries. 
The CAA loss encourages both shapes and colors to concentrate on the correct regions instead of background areas without pertinent concepts.
This mechanism improves the accurate capture of the intended attributes.

\noindent{\bf Training scheme.} \ourmethod ensures that the color token $\colortoken$ effectively extracts color attributes from a given image while disentangling them from the shapes. 
As a secondary benefit, it also allows the shape token $\shapetoken$ to learn novel shapes that are not present in the T2I diffusion models. In our experiments, we demonstrate that disentangling both colors and shapes leads to significantly better performance than disentangling color attributes alone. For successful disentanglement, each color should have at least two shapes, and vice versa. 

\section{Experiments}
\label{sec:experiments}

\subsection{Experimental setup}
\label{subsec:exp_setup}

\noindent{\bf Dataset.} 
Colors of objects in real-world images are influenced by various scene factors like illuminant color, viewing angle, and shadows. To assess learning performance of \ourmethod, we develop an automatic color synthesizer capable of generating basic $2$D and $3$D shapes with specified colors and shapes using RGB triplets. For the $2$D dataset, we incorporate the shapes circle, square, hexagon, and triangle. For the $3$D dataset, we curate a small collection comprising $200$ images of $3$D shapes with attributes such as colors, textures, reflectance, and lighting. Our blender-designed dataset encompasses five $3$D shapes: sphere, cylinder, hexagon, cube, and cone. For color prompt learning, we create two subsets: coarse-grained (red, green, blue, yellow) and fine-grained ($18$ colors related to less common color names, including 'salmon','beige', etc). Users can synthesize shapes in any desired color using our dataset synthesizer, given the RGB triplet. Further details regarding our dataset are available in Section~\ref{sec:dataset_details}.

\minisection{Evaluation metrics.}
Quantitatively analyzing colors in T2I generation poses notable challenges including lighting variation, reflections, and illuminant temperature which can lead to inaccuracies in the analysis. To address these challenges effectively, we compute the following metrics: (i) \textit{Euclidean Distance in CIE Lab color space ($\Delta E$, $\Delta E_{Ch}$ when luminance is removed)}  --- to analyze perceptual uniformity between the generated and given color, (iii) \textit{Mean Angular Error (MAE) in sRGB} --- to understand the color deviation in terms of chromaticity, and (iii) \textit{Mean Angular Error (MAE) in Hue} --- to analyze the difference between given and generated color irrespective of brightness and saturation. For each comparison method, we generated images using 10 prompts, each with 20 random seeds. After image generation, we use the Segment-Anything model~\cite{kirillov2023segment} to derive object masks, delineating regions for the computation of evaluation metrics. Lastly, we extract our generated object from the image using the mask and compute the aforementioned metrics based on the user-provided color. Additional details are provided in Fig.~\ref{fig:gen_mask} in the supplementary material.

\noindent{\bf Implementation details.}
We demonstrate our method \ourmethod in various experiments based on the open-source T2I model Stable Diffusion~\cite{Rombach_2022_CVPR_stablediffusion} following previous methods~\cite{ruiz2022dreambooth,kumari2022customdiffusion,textual_inversion}. 
We train \ourmethod  with batch size of 2 and a learning rate of $10^{-5}$. For the coarse-color learning, we train the model for 1500 steps. Whereas, we increase training steps to 6000 steps for fine-grained color learning.  All experiments are done on A40 GPUs. 

\noindent{\bf Comparison methods.}
Firstly, we evaluate Stable Diffusion and Rich-Text~\cite{ge2023richtext} methods to analyze the color generation from RGB values, and specific color names in the text prompts. Secondly, we analyze seminal personalization methods, including Textual Inversion, Dreambooth and Custom Diffusion. For Textual Inversion, $\colortokeni$ and $\shapetokeni$ --- two new learnable tokens --- are optimized to learn the color and shape, respectively. Whereas, for Dreambooth, $\colortokeni$ and $\shapetokeni$ are initialized with the existing rare tokens, which are optimized along with all the parameters in the diffusion model. We also compare with the Custom Diffusion baseline, which optimizes the $\colortokeni$ and $\shapetokeni$ for color and shape, along with key and value projection matrices in the diffusion models. Following~\cite{kumari2022customdiffusion}, TI and DB are optimized for 4000 steps, while CD is optimized for 1500 steps to perform the coarse-grained learning task.
More details on the compared methods are included in the supplementary material.

\subsection{Qualitative Comparisons}

\begin{figure}[t]
  \centering
  \includegraphics[width=0.999\textwidth]{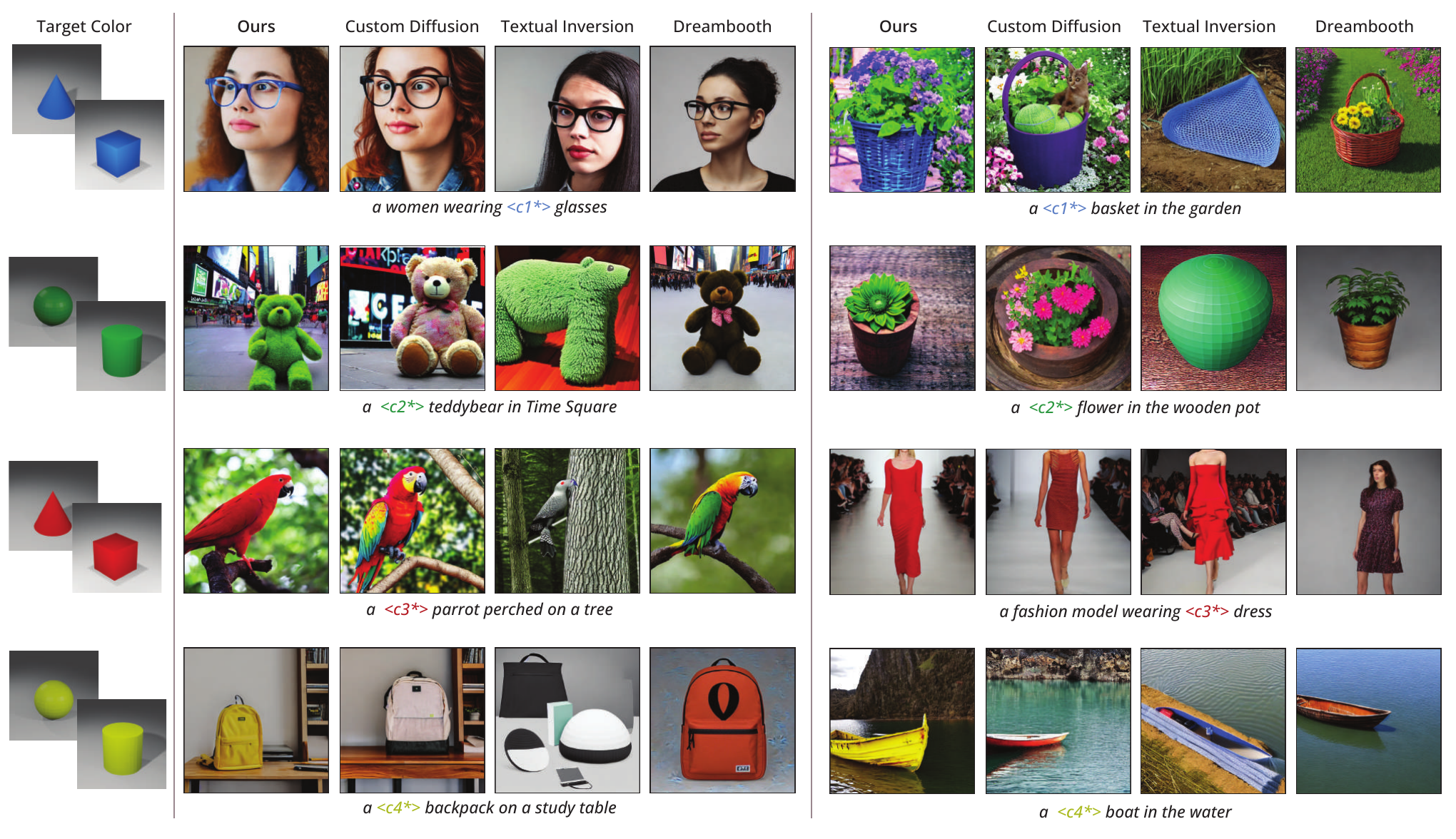}
  \vspace{-6mm}
  \caption{Qualitative results on the coarse-grained \textit{color prompt learning} task compared with other T2I model adaptation methods including CustomDiffusion~\cite{kumari2022customdiffusion}, DreamBooth~\cite{ruiz2022dreambooth} and Textual Inversion~\cite{textual_inversion}.
  }
  \vspace{-4mm}
  \label{fig:coarse_expr}
\end{figure}

\begin{figure}[t]
  \centering
  \includegraphics[width=0.9999\textwidth]{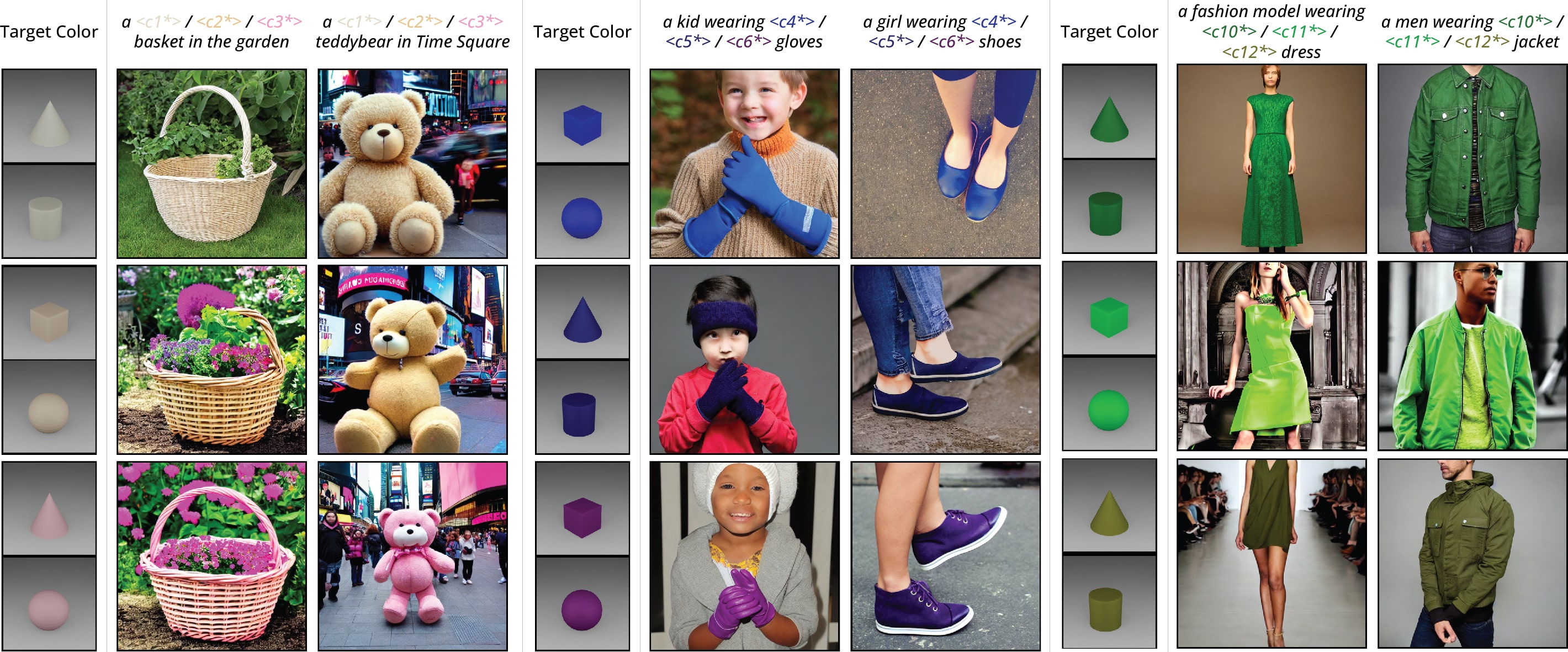}
  \vspace{-4mm}
  \caption{Qualitative results of fine-grained color learning. From customizing backgrounds, dresses, and shoes to eyes, our method \ourmethod can generate high-quality variations in fine-grained color concepts.
  }
  \vspace{-1mm}
  \label{fig:fine_expr}
\end{figure}

In Fig.~\ref{fig:coarse_expr} and Fig.~\ref{fig:fine_expr},  we show the performance of \ourmethod applied in both situations of color prompt learning: coarse-grained and fine-grained color concepts. 

\begin{figure}[t]
  \centering
  \includegraphics[width=0.999\textwidth]{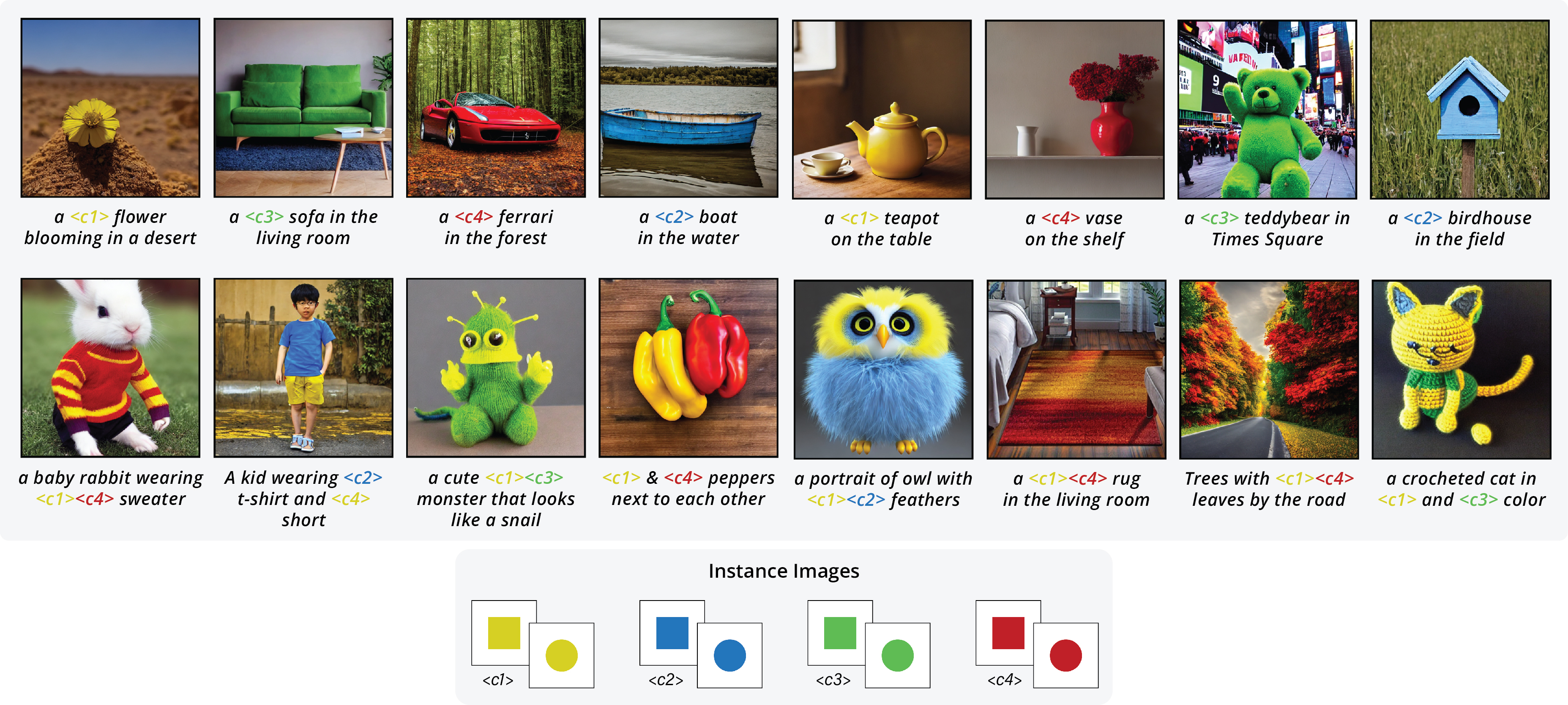}
  \vspace{-5mm}
  \caption{Qualitative results of \ourmethod in single color and multi-color compositions.}
  \vspace{-5mm}
  \label{fig:cp_results}
\end{figure}

\noindent{\bf Coarse-Grained color concepts.} First, we conduct experiments over coarse colors including red, green, blue, and yellow to analyze the learning of color prompts from given colored shapes and their transferability to real concept compositions. To evaluate if \ourmethod is correctly disentangling the colors from shapes, 
we optimize $\colortokeni$ and $\shapetokeni$ to learn the color and shape in the training set. The results are illustrated in Fig.~\ref{fig:cp_results} which show that \ourmethod can efficiently generate the concepts in the user-provided colors. In particular, our method can generate precise colors for both the single and multiple concepts ranging from objects in complex scenes to intricate attributes such as eye color of the cat, wings of the parrot and more. 

In the next step, we analyze the color transferability of \ourmethod to real-world scenarios and compare with Custom Diffusion, Textual Inversion, and Dreambooth. From Fig.~\ref{fig:coarse_expr}, it can be observed that \ourmethod generates more realistic concepts as compared to the existing new-concept learning methods. Unlike Textual Inversion and Dreambooth, which tend to ignore the target prompt, \ourmethod ensures the high quality color transferability in terms of consistency and fidelity. Moreover, we did not observe any evidence of \textit{overfitting} in the generated results. We hypothesize that this is because colors are abstract in nature in contrast to the learning of real objects from images---which may align with the prior knowledge of stable diffusion.

\noindent{\bf Fine-Grained color concepts.} Next, to illustrate the efficacy of \ourmethod, we design the harder task as composing with fine-grained concepts. Here we leverage our fine-grained color learning dataset which contains several variants of colors such as blue, cyan, navy, indigo (see Section~\ref{sec:dataset_details} for details) to learn fine-grained colors and illustrated the results in Fig.~\ref{fig:fine_expr}. As can be seen, it is evident that \ourmethod efficiently distinguishes the fine-grained colors and generates highly detailed concepts aligning with the given text prompts. Other than high quality image generation, \ourmethod also demonstrates efficient customization of various elements ranging from personalized dressing concepts (such as clothes, footwear, gloves, glasses) to toys/objects in different scenarios.

\begin{figure}[t]
  \centering
  \includegraphics[width=0.999\textwidth]{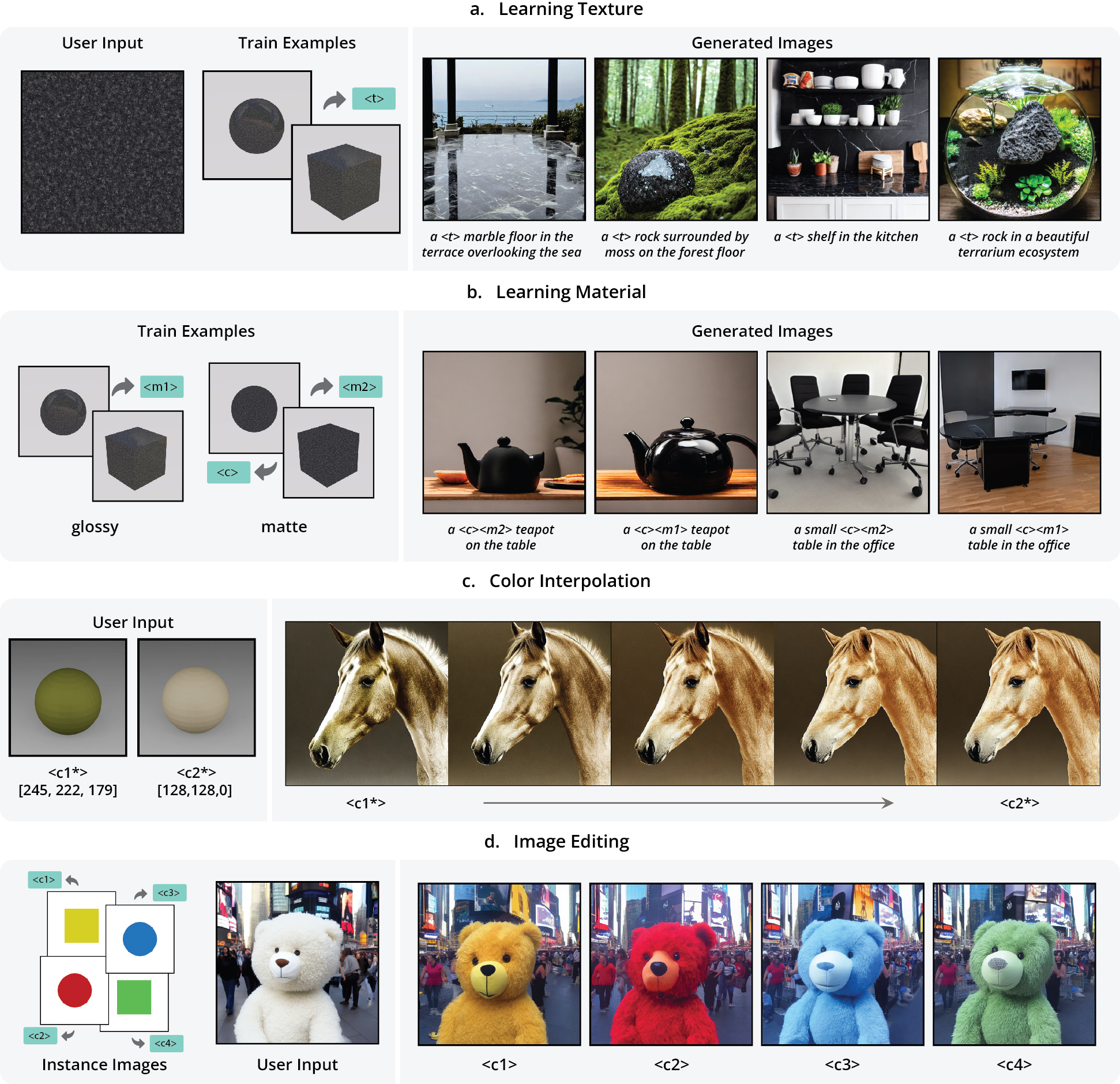}
  \vspace{-4mm}
  \caption{Demonstrating generalization of \ourmethod to (a) Texture Learning, (b) Material Learning, (c) Color Interpolation, and (d) Image Editing.
  }
  \vspace{-3mm}
  \label{fig:generalization}
\end{figure}

\noindent{\bf \ourmethod generalizability.}
Other than learning colors, \ourmethod can be extended to learning texture and material from the user-provided input image as shown in Fig.~\ref{fig:generalization}. Similar to the color prompt learning, firstly, the given 2D texture image is mapped on to the 3D shapes such as sphere or cube, using our dataset synthesizer. As a result, we get the train examples for textures (Fig.~\ref{fig:generalization}a) and materials (Fig.~\ref{fig:generalization}b). Secondly, we denote the target texture and material as $\texturetoken$ and $\materialtoken$, respectively. In the next step, to learn novel texture and material token embeddings ($\tokenset^{\texturetoken}$, $\tokenset^{\materialtoken}$), we randomly sample an instance image from our small training set, which depict our target material and texture in various shapes. We directly optimize new tokens ($\tokenset^{\texturetoken}$, $\tokenset^{\materialtoken}$) as discussed in section~\ref{subsec:cpl}.

\minisection{Color token interpolation.}
We also included an initial linear interpolation result between two color tokens, which shows that already \ourmethod can represent the colors continuously between the learned color prompts (Fig.~\ref{fig:generalization}c). This can avoid the training for new colors. In Section~\ref{subsec:color_interpolation}, we also include results with interpolation between several colors. 

\minisection{Image editing.} For text-guided image editing, we follow the P2P~\cite{hertz2022prompt} approach by swapping the cross-attention maps during the inference stage. The corresponding image editing results are shown in Fig.~\ref{fig:generalization}d, where we successfully modify the color of the teddybear into our learned colors. More examples are shown in Fig.~\ref{fig:image_editing} the supplementary.

\subsection{Quantitative Analysis}
We compare \ourmethod with: \textbf{(i) T2I generation} --- Stable Diffusion v1.4, and Rich-Text~\cite{ge2023richtext} based on Stable Diffusion(SD), and \textbf{(ii) personalization methods} --- DreamBooth(DB), Textual Inversion(TI) and Custom Diffusion(CD). For each method, we generated images, and extract the mask of the object--- discussed in section \ref{subsec:exp_setup}. The results are summarized in Table~\ref{Fig:Quantitative comparison}, where they are provided as the Median for all the images. Percentages in MAE metrics denote the percentage of pixels inside the mask used for the computation (selecting those closest to the ground truth). 
This table clearly show the superiority of \ourmethod. Particularly, \ourmethod achieved notably lower $\Delta E$ error in CIE Lab color space as compared to the existing methods which indicates that \ourmethod generates perceptually better colors. In addition, \ourmethod also achieved comparatively much lower mean angular error in both sRGB and Hue, which signifies a higher degree of color accuracy in terms of chromaticity and hue in our generated images. To demonstrate the adaptability of our method, we integrated \ourmethod with DreamBooth to analyze its performance. The results show that \ourmethod significantly enhances DreamBooth's performance, particularly in terms of color accuracy in the generated images. Some of the examples are demonstrated in supplementary (see Fig.~\ref{fig:fine_grained_gen_}).

\begin{table}[t]
    \centering
    \begin{minipage}[t]{0.52\textwidth}
        \centering
        \caption{Quantitative comparison with baselines over various evaluation metrics. All numbers are the smaller the better ($\downarrow$). Best result is in bold, second best is underlined.
        }\label{Fig:Quantitative comparison}
        \resizebox{\textwidth}{!}{
        \begin{tabular}{ c | c | c | ccc | ccc | c }
        \toprule
        \multirow{2}{*}{{Method}}  &  \multirow{2}{*}{{$\Delta E$}} 	&   \multirow{2}{*}{{$\Delta E_{ch}$}} 	 & \multicolumn{3}{c|}{MAE (rgb)}	&  \multicolumn{3}{c|}{MAE (Hue)}  &  Time \\
        
        & & & 10\% & 50\% & 100\% & 10\% & 50\% & 100\% & (min) \\
        \midrule
        SD~\cite{Rombach_2022_CVPR_stablediffusion} & 47.45 & 41.55 & 12.89 & 20.04 & 26.93 & 30.17 & 54.14 & 86.38 & - \\
        Rich-Text~\cite{ge2023richtext} & 36.62 & 32.48 & 9.91 &  13.29 & 18.53 & 50.55 & 72.77 & 93.51 & - \\
        \midrule 
        TI~\cite{textual_inversion} & 48.98 & 44.29 & 15.22 & 19.51 & 23.90 & 52.66 & 69.35 & 90.88 & 118  \\
        DB~\cite{ruiz2022dreambooth} & 50.71 & 46.29 & 14.75 & 19.30 & 23.70 & 47.12 & 67.13 & 88.72 & 56  \\
        CD~\cite{kumari2022customdiffusion} & 48.47 & 42.23 & 13.43 & 17.93 & 22.43 & 31.63 & 55.07 & 78.43 & 24  \\
        \midrule
        \ourmethod (3D) & \underline{21.39} &	\underline{16.51}	& \textbf{4.36}	& \textbf{7.76} &	\textbf{12.08} &	\textbf{2.63} & \textbf{6.47} & \textbf{21.35} & \multirow{2}{*}{{\textbf{19}}} \\
        \ourmethod (2D) & \textbf{20.45} &	\textbf{15.29}	& \underline{4.83}	& \underline{7.88} &	\underline{12.13} &	\underline{3.18} & \underline{7.43} & \underline{21.46} & \\
        \bottomrule
        \end{tabular}
        }
    \end{minipage}\hfill
    \begin{minipage}[t]{0.473\textwidth}
        \centering
        \caption{Ablation study over hyperparameters $\lambda$ on \ourmethod (3D). All numbers are smaller the better ($\downarrow$). Best result is in bold.}\label{tab:ablation}
        \resizebox{\textwidth}{!}{
        \begin{tabular}{ c | c | c | ccc | ccc }
        \toprule
        \multirow{2}{*}{{$\lambda$}} & \multirow{2}{*}{{$\Delta E$}} 	&   \multirow{2}{*}{{$\Delta E_{Ch}$}} 	 & \multicolumn{3}{c|}{MAE (rgb)}	&  \multicolumn{3}{c}{MAE (Hue)} \\
        
        & & & 10\% & 50\% & 100\% & 10\% & 50\% & 100\% \\
        
        \midrule
        0.0 (CD) & 48.47 & 42.23 & 13.43 & 17.93 & 22.43 & 31.63 & 55.07 & 78.43 \\
        0.1  & 22.23 & 16.86 & 5.13 & 8.63 & 12.75 & 3.48 & 10.54 & 36.36 \\
        0.2  & \textbf{21.39} &	\textbf{16.51}	& \textbf{4.36}	& \textbf{7.76} &	\textbf{12.08} &	\textbf{2.63} & \textbf{6.47} & \textbf{21.35} \\
        0.4  & 23.37 & 17.10 & 4.91 & 8.46 & 12.77 & 3.87 & 8.65 & 24.94 \\
        0.6  & 23.53 & 16.75 & 4.97 & 8.48 & 13.25 & 2.89 & 8.96 & 28.39 \\
        0.8  & 23.79 & 17.01 & 4.98 & 8.57 & 13.50 & 4.06 & 13.80 & 33.54  \\
        1.0  & 24.43 & 18.64 & 5.03 & 8.69 & 13.77 & 4.27 & 10.48 & 34.35 \\

        \bottomrule
        \end{tabular}
        }
    \end{minipage}
\end{table}


\begin{figure}[htbp]
    \centering
    \begin{subfigure}[b]{0.60\textwidth}
        \centering
        \includegraphics[width=0.955\textwidth]{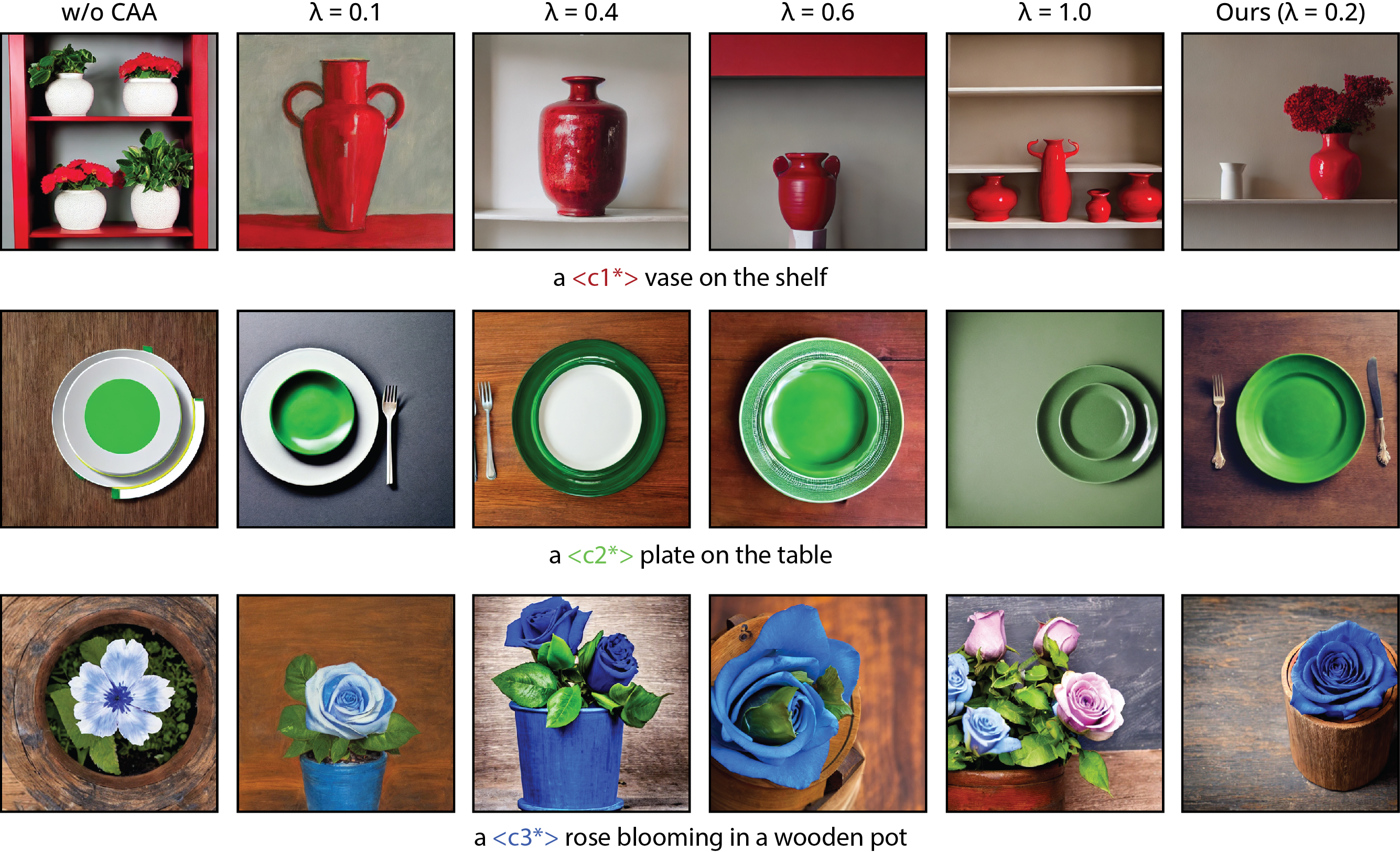}
        \caption{Ablation Study}
        \label{fig:ablation}
    \end{subfigure}
    \hfill
    \begin{subfigure}[b]{0.39\textwidth}
        \centering
        \includegraphics[width=\textwidth]{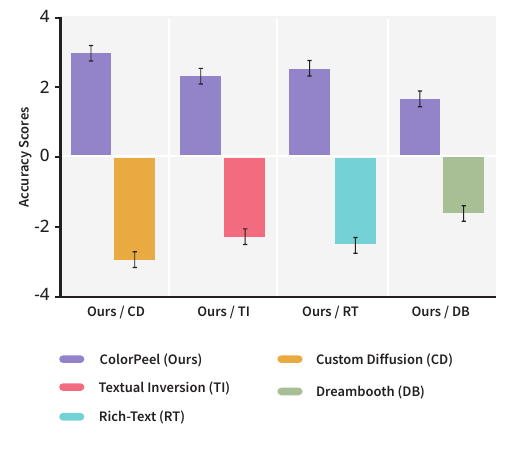}
        \caption{Human Study}
        \label{fig:User_results}
    \end{subfigure}
    \vspace{-4mm}
    \caption{Illustration of ablation and psychophysical (human) study. \textbf{(a)} We remove cross attention alignment loss, and scale lambda to demonstrate the effect on color fidelity and transferability in image generation. As can be seen, the model fails to disentangle the shape and color when our proposed cross attention alignment (CAA) loss  is removed. \textbf{(b)} Thurstone case V results of our user's study. Values are z-scores. Error bars represent 95\% confidence intervals ---see \cite{montag2006}. Our method is statistical significantly better than existing methods ---CD, DB, Rich-text, and TI.}
    \label{fig:both_images}
    \vspace{-2mm}
\end{figure}


\minisection{User study.}
We conducted a user study with 15 participants to perceptually evaluate our results, comparing \ourmethod against TI, Rich-Text, DB, and CD. The experiment was conducted in a controlled lab environment to ensure the reliability of our study. All observers were tested for correct color vision using the Ishihara test. The experiment employed a two-alternative forced choice (2AFC) method. Observers were presented with three images on a monitor set to sRGB. The central image represents the desired color. To the left and right, we displayed the results of the given prompt by our method and one of the competing methods, with the order randomized. We tested $10$ different prompts and $4$ different colors (red, green, blue, and yellow), consistent with the quantitative analysis. We analyzed the results by comparing \ourmethod to each of the others using the Thurstone Case V Law of Comparative Judgment model \cite{thurstone2017}. This method provided us with z-scores and a $95\%$ confidence interval, calculated using the method proposed in \cite{montag2006}. The results are presented in Fig.~\ref{fig:User_results}. We observe that our approach \ourmethod is statistically significantly better than any of the competing $4$ algorithms. These findings underscore the effectiveness of \ourmethod in generating more realistic and accurate colors given an RGB triplet.

\minisection{Ablation Study.}
In Fig.~\ref{fig:ablation}, we conduct ablation studies over various factors. Here we analyze the disentanglement and transferability of colors to the real objects. We note that by removing the cross attention alignment loss, the model struggles to disentangle the color from the shape and fails to preserve the identity. Moreover, we can see that the model generates inconsistent colors when $\lambda$ in CAA is scaled up or down in the cross-attention alignment loss, which is also reflected in Table~\ref{tab:ablation}. Note that with $\lambda=0.0$, \ourmethod degrades to CD, that shows attention leakage (see Fig. {\ref{fig:supp_attn_maps}) and failures in color generations (see Fig.~\ref{fig:baselines}, Fig.~\ref{fig:baselines_lc}). We also show $\lambda=0.0$ results in Fig.~\ref{fig:ablation} and Table~\ref{tab:ablation}. To further analyze the role of CAA in disentanglement of shape from color, we reproduce the instance prompt without CAA, and note that the model tends to ignore the target text prompt, replicates the instance images, while also failing to transfer the colors to other shapes (see Fig.~\ref{fig:ablations_expanded} in supplementary). Attention leakage has been studied in T2I generation ~\cite{rassin2024linguistic}, but not explored in T2I personalization. This problem is address by our CAA loss.

\section{Conclusion}
\label{sec:conclusion}

Text-to-Image (T2I) diffusion models have encountered challenges when generating specific object colors using linguistic color names, referred to as the \textit{color prompt learning} task. 
Existing T2I models and adaptation methods struggle to tackle this challenge effectively. In this paper, we propose a method called \ourmethod to learn specific \textit{color prompts} tailored to user-selected colors. We achieve this by generating basic geometric objects in the target color and employing disentanglement to \textit{peel off} the color from the shapes. These tailored prompts are then used to generate objects with the desired colors precisely.
\ourmethod enhances the precision of color generation within the T2I framework, and our experimental results demonstrate its effectiveness.
Moreover, we showcase its utility in real-world applications such as recoloring objects within images and its generalizability in learning new textures, material, etc.
In summary, our research contributes to improving the precision and versatility of T2I models, opening up new possibilities for creative applications and design tasks. By addressing the challenge of precise color specification, \ourmethod advances T2I technology and holds promise for various practical applications.

\minisection{Acknowledgments.} We acknowledge  projects TED2021-132513B-I00, PID2021-128178OB-I00 and PID2022-143257NB-I00, financed by MCIN / AEI / 10.13039 / 501100011033 and FSE+ by the European Union NextGenerationEU/PRTR, and by ERDF A Way of Making Europa, the  Departament de Recerca i Universitats from Generalitat de Catalunya with reference 2021SGR01499, and the Generalitat de Catalunya CERCA Program.


\bibliographystyle{splncs04}
\bibliography{longstrings,main}

\begin{thebibliography}{10}
\providecommand{\url}[1]{\texttt{#1}}
\providecommand{\urlprefix}{URL }
\providecommand{\doi}[1]{https://doi.org/#1}

\bibitem{avrahami2023breakascene}
Avrahami, O., Aberman, K., Fried, O., Cohen-Or, D., Lischinski, D.: Break-a-scene: Extracting multiple concepts from a single image. SIGGRAPH Asia 2023  (2023)

\bibitem{basu2023editval}
Basu, S., Saberi, M., Bhardwaj, S., Chegini, A.M., Massiceti, D., Sanjabi, M., Hu, S.X., Feizi, S.: Editval: Benchmarking diffusion based text-guided image editing methods. arXiv preprint arXiv:2310.02426  (2023)

\bibitem{berlin1991basic}
Berlin, B., Kay, P.: Basic color terms: Their universality and evolution. Univ of California Press (1991)

\bibitem{brooks2022instructpix2pix}
Brooks, T., Holynski, A., Efros, A.A.: Instructpix2pix: Learning to follow image editing instructions. In: CVPR (2023)

\bibitem{chang2023muse}
Chang, H., Zhang, H., Barber, J., Maschinot, A., Lezama, J., Jiang, L., Yang, M.H., Murphy, K., Freeman, W.T., Rubinstein, M., et~al.: Muse: Text-to-image generation via masked generative transformers. ICML  (2023)

\bibitem{chen2023fec}
Chen, S., Huang, J.: Fec: Three finetuning-free methods to enhance consistency for real image editing. arXiv preprint arXiv:2309.14934  (2023)

\bibitem{chen2023suti}
Chen, W., Hu, H., Li, Y., Rui, N., Jia, X., Chang, M.W., Cohen, W.W.: Subject-driven text-to-image generation via apprenticeship learning. arXiv preprint arXiv:2304.00186  (2023)

\bibitem{couairon2023diffedit}
Couairon, G., Verbeek, J., Schwenk, H., Cord, M.: Diffedit: Diffusion-based semantic image editing with mask guidance. In: The Eleventh International Conference on Learning Representations (2023), \url{https://openreview.net/forum?id=3lge0p5o-M-}

\bibitem{daras2022multiresolution}
Daras, G., Dimakis, A.: Multiresolution textual inversion. In: NeurIPS 2022 Workshop on Score-Based Methods (2022)

\bibitem{dong2022dreamartist}
Dong, Z., Wei, P., Lin, L.: Dreamartist: Towards controllable one-shot text-to-image generation via contrastive prompt-tuning. arXiv preprint arXiv:2211.11337  (2022)

\bibitem{gafni2022make}
Gafni, O., Polyak, A., Ashual, O., Sheynin, S., Parikh, D., Taigman, Y.: Make-a-scene: Scene-based text-to-image generation with human priors. In: ECCV. pp. 89--106. Springer (2022)

\bibitem{textual_inversion}
Gal, R., Alaluf, Y., Atzmon, Y., Patashnik, O., Bermano, A.H., Chechik, G., Cohen-Or, D.: An image is worth one word: Personalizing text-to-image generation using textual inversion. ICLR  (2023)

\bibitem{gal2023designing}
Gal, R., Arar, M., Atzmon, Y., Bermano, A.H., Chechik, G., Cohen-Or, D.: Designing an encoder for fast personalization of text-to-image models. arXiv preprint arXiv:2302.12228  (2023)

\bibitem{ge2023richtext}
Ge, S., Park, T., Zhu, J.Y., Huang, J.B.: Expressive text-to-image generation with rich text. In: Proceedings of the IEEE/CVF International Conference on Computer Vision. pp. 7545--7556 (2023)

\bibitem{hiper2023}
Han, I., Yang, S., Kwon, T., Ye, J.C.: Highly personalized text embedding for image manipulation by stable diffusion. arXiv preprint arXiv:2303.08767  (2023)

\bibitem{han2023svdiff}
Han, L., Li, Y., Zhang, H., Milanfar, P., Metaxas, D., Yang, F.: Svdiff: Compact parameter space for diffusion fine-tuning. ICCV  (2023)

\bibitem{han2023ProxNPI}
Han, L., Wen, S., Chen, Q., Zhang, Z., Song, K., Ren, M., Gao, R., Chen, Y., Liu, D., Zhangli, Q., et~al.: Improving negative-prompt inversion via proximal guidance. arXiv preprint arXiv:2306.05414  (2023)

\bibitem{hertz2023delta_DDS}
Hertz, A., Aberman, K., Cohen-Or, D.: Delta denoising score. arXiv preprint arXiv:2304.07090  (2023)

\bibitem{hertz2022prompt}
Hertz, A., Mokady, R., Tenenbaum, J., Aberman, K., Pritch, Y., Cohen-Or, D.: Prompt-to-prompt image editing with cross attention control. ICLR  (2023)

\bibitem{ho2022imagen}
Ho, J., Chan, W., Saharia, C., Whang, J., Gao, R., Gritsenko, A., Kingma, D.P., Poole, B., Norouzi, M., Fleet, D.J., et~al.: Imagen video: High definition video generation with diffusion models. arXiv preprint arXiv:2210.02303  (2022)

\bibitem{ho2022classifier}
Ho, J., Salimans, T.: Classifier-free diffusion guidance. NeurIPS 2021 Workshop on Deep Generative Models and Downstream Applications  (2022)

\bibitem{hong2022sag}
Hong, S., Lee, G., Jang, W., Kim, S.: Improving sample quality of diffusion models using self-attention guidance. ICCV  (2023)

\bibitem{huang2024diffusion_image_edit_survey}
Huang, Y., Huang, J., Liu, Y., Yan, M., Lv, J., Liu, J., Xiong, W., Zhang, H., Chen, S., Cao, L.: Diffusion model-based image editing: A survey. arXiv preprint arXiv:2402.17525  (2024)

\bibitem{direct_inversion_2023}
Ju, X., Zeng, A., Bian, Y., Liu, S., Xu, Q.: Direct inversion: Boosting diffusion-based editing with 3 lines of code. arXiv preprint arXiv:2310.01506  (2023)

\bibitem{kawar2022imagic}
Kawar, B., Zada, S., Lang, O., Tov, O., Chang, H., Dekel, T., Mosseri, I., Irani, M.: Imagic: Text-based real image editing with diffusion models. CVPR  (2023)

\bibitem{kirillov2023segment_anything_sam}
Kirillov, A., Mintun, E., Ravi, N., Mao, H., Rolland, C., Gustafson, L., Xiao, T., Whitehead, S., Berg, A.C., Lo, W.Y., Doll{\'a}r, P., Girshick, R.: Segment anything. ICCV  (2023)

\bibitem{kirillov2023segment}
Kirillov, A., Mintun, E., Ravi, N., Mao, H., Rolland, C., Gustafson, L., Xiao, T., Whitehead, S., Berg, A.C., Lo, W.Y., et~al.: Segment anything. arXiv preprint arXiv:2304.02643  (2023)

\bibitem{kumari2022customdiffusion}
Kumari, N., Zhang, B., Zhang, R., Shechtman, E., Zhu, J.Y.: Multi-concept customization of text-to-image diffusion. CVPR  (2023)

\bibitem{li2023stylediffusion}
Li, S., van~de Weijer, J., Hu, T., Khan, F.S., Hou, Q., Wang, Y., Yang, J.: Stylediffusion: Prompt-embedding inversion for text-based editing (2023)

\bibitem{Cones2023}
Liu, Z., Feng, R., Zhu, K., Zhang, Y., Zheng, K., Liu, Y., Zhao, D., Zhou, J., Cao, Y.: Cones: Concept neurons in diffusion models for customized generation. ICML  (2023)

\bibitem{lopes2023material}
Lopes, I., Pizzati, F., de~Charette, R.: Material palette: Extraction of materials from a single image. arXiv preprint arXiv:2311.17060  (2023)

\bibitem{meng2022sdedit}
Meng, C., He, Y., Song, Y., Song, J., Wu, J., Zhu, J.Y., Ermon, S.: {SDE}dit: Guided image synthesis and editing with stochastic differential equations. In: International Conference on Learning Representations (2022), \url{https://openreview.net/forum?id=aBsCjcPu_tE}

\bibitem{miyake2023NPI}
Miyake, D., Iohara, A., Saito, Y., Tanaka, T.: Negative-prompt inversion: Fast image inversion for editing with text-guided diffusion models. arXiv preprint arXiv:2305.16807  (2023)

\bibitem{mokady2022null}
Mokady, R., Hertz, A., Aberman, K., Pritch, Y., Cohen-Or, D.: Null-text inversion for editing real images using guided diffusion models. CVPR  (2023)

\bibitem{montag2006}
Montag, E.D.: Empirical formula for creating error bars for the method of paired comparison. J. Elec. Imag.  \textbf{15}(1),  010502--010502 (2006)

\bibitem{motamed2023lego}
Motamed, S., Paudel, D.P., Van~Gool, L.: Lego: Learning to disentangle and invert concepts beyond object appearance in text-to-image diffusion models. arXiv preprint arXiv:2311.13833  (2023)

\bibitem{parmar2023zero}
Parmar, G., Singh, K.K., Zhang, R., Li, Y., Lu, J., Zhu, J.Y.: Zero-shot image-to-image translation. Proceedings of the ACM SIGGRAPH Conference on Computer Graphics  (2023)

\bibitem{podell2023sdxl}
Podell, D., English, Z., Lacey, K., Blattmann, A., Dockhorn, T., M{\"u}ller, J., Penna, J., Rombach, R.: Sdxl: improving latent diffusion models for high-resolution image synthesis. arXiv preprint arXiv:2307.01952  (2023)

\bibitem{ramesh2022dalle2}
Ramesh, A., Dhariwal, P., Nichol, A., Chu, C., Chen, M.: Hierarchical text-conditional image generation with clip latents. arXiv preprint arXiv:2204.06125  (2022)

\bibitem{ramesh2021zero}
Ramesh, A., Pavlov, M., Goh, G., Gray, S., Voss, C., Radford, A., Chen, M., Sutskever, I.: Zero-shot text-to-image generation. In: International Conference on Machine Learning. pp. 8821--8831. PMLR (2021)

\bibitem{rassin2024linguistic}
Rassin, R., Hirsch, E., Glickman, D., Ravfogel, S., Goldberg, Y., Chechik, G.: Linguistic binding in diffusion models: Enhancing attribute correspondence through attention map alignment. Advances in Neural Information Processing Systems  \textbf{36} (2024)

\bibitem{Rombach_2022_CVPR_stablediffusion}
Rombach, R., Blattmann, A., Lorenz, D., Esser, P., Ommer, B.: High-resolution image synthesis with latent diffusion models. In: Proceedings of the IEEE/CVF Conference on Computer Vision and Pattern Recognition (CVPR). pp. 10684--10695 (June 2022)

\bibitem{ronneberger2015unet}
Ronneberger, O., Fischer, P., Brox, T.: U-net: Convolutional networks for biomedical image segmentation. In: Medical Image Computing and Computer-Assisted Intervention--MICCAI 2015: 18th International Conference, Munich, Germany, October 5-9, 2015, Proceedings, Part III 18. pp. 234--241. Springer (2015)

\bibitem{ruiz2022dreambooth}
Ruiz, N., Li, Y., Jampani, V., Pritch, Y., Rubinstein, M., Aberman, K.: Dreambooth: Fine tuning text-to-image diffusion models for subject-driven generation. CVPR  (2023)

\bibitem{saharia2022imagen}
Saharia, C., Chan, W., Saxena, S., Li, L., Whang, J., Denton, E., Ghasemipour, S.K.S., Ayan, B.K., Mahdavi, S.S., Lopes, R.G., et~al.: Photorealistic text-to-image diffusion models with deep language understanding. Advances in Neural Information Processing Systems  (2022)

\bibitem{shi2023instantbooth}
Shi, J., Xiong, W., Lin, Z., Jung, H.J.: Instantbooth: Personalized text-to-image generation without test-time finetuning. arXiv preprint arXiv:2304.03411  (2023)

\bibitem{deepfloyd}
Shonenkov, A., Konstantinov, M., Bakshandaeva, D., Schuhmann, C., Ivanova, K., Klokova, N.: Deepfloyd-if. \url{https://github.com/deep-floyd/IF} (2023)

\bibitem{singh2006impact}
Singh, S.: Impact of color on marketing. Management decision  \textbf{44}(6),  783--789 (2006)

\bibitem{tang2023iterinv}
Tang, C., Wang, K., van~de Weijer, J.: Iterinv: Iterative inversion for pixel-level t2i models. Neurips 2023 workshop on Diffusion Models  (2023)

\bibitem{tang2024locinv}
Tang, C., Wang, K., Yang, F., van~de Weijer, J.: Locinv: Localization-aware inversion for text-guided image editing. CVPR 2024 AI4CC workshops  (2024)

\bibitem{thurstone2017}
Thurstone, L.L.: A law of comparative judgment. In: Scaling, pp. 81--92. Routledge (1927)

\bibitem{tumanyan2022plug}
Tumanyan, N., Geyer, M., Bagon, S., Dekel, T.: Plug-and-play diffusion features for text-driven image-to-image translation. CVPR  (2023)

\bibitem{van2009learning}
Van De~Weijer, J., Schmid, C., Verbeek, J., Larlus, D.: Learning color names for real-world applications. IEEE TIP  \textbf{18}(7),  1512--1523 (2009)

\bibitem{vinker2023concept_decomposition}
Vinker, Y., Voynov, A., Cohen-Or, D., Shamir, A.: Concept decomposition for visual exploration and inspiration. SIGGRAPH Asia 2023  (2023)

\bibitem{voynov2023ETI}
Voynov, A., Chu, Q., Cohen-Or, D., Aberman, K.: $p+$: Extended textual conditioning in text-to-image generation. arXiv preprint arXiv:2303.09522  (2023)

\bibitem{kai2023DPL}
Wang, K., Yang, F., Yang, S., Butt, M.A., van~de Weijer, J.: Dynamic prompt learning: Addressing cross-attention leakage for text-based image editing. Advances in Neural Information Processing Systems  (2023)

\bibitem{yeh2024texturedreamer}
Yeh, Y.Y., Huang, J.B., Kim, C., Xiao, L., Nguyen-Phuoc, T., Khan, N., Zhang, C., Chandraker, M., Marshall, C.S., Dong, Z., et~al.: Texturedreamer: Image-guided texture synthesis through geometry-aware diffusion. arXiv preprint arXiv:2401.09416  (2024)

\bibitem{zhang2023forgedit}
Zhang, S., Xiao, S., Huang, W.: Forgedit: Text guided image editing via learning and forgetting. arXiv preprint arXiv:2309.10556  (2023)

\bibitem{zhang2022sine}
Zhang, Z., Han, L., Ghosh, A., Metaxas, D., Ren, J.: Sine: Single image editing with text-to-image diffusion models. CVPR  (2023)

\end{thebibliography}

\clearpage

\appendix
\section*{Supplementary Material}

\section{Boarder Impacts and Limitations}

\minisection{Boarder Impacts.}
The application of T2I models in image editing and generation offers extensive potential for diverse downstream applications, enabling the adaptation of images to different contexts. In particular, synthesizing objects in the precise colors has diverse applications, however, it is a challenging task for diffusion models. Our \ourmethod can help the users to customize their desiring objects in the precise colors, given an RGB triplet or color picker, resulting in significant time and resource savings. Notably, current methods have inherent limitations, as discussed in this paper. However, our model can serve as an intermediary solution while offering valuable insights for further advancements.

\minisection{Limitations.}
\label{sec:limitation}
While our method \ourmethod can achieve high-fidelity personalization diffusion models with \textit{color prompt learning}, it is not free of limitations.
Firstly, colors encompass a vast spectrum that extends to countless combinations of hues and shades. Therefore, it is hard to learn such wider range of color concepts with current approach due to inherent limitations of T2I personalization methods such as identity preservation. In future research, we aim to explore learning a color grid of tokens, allowing users to interpolate the color combinations, to avoid training for new colors. Secondly, in the current study, we do not explicitly decompose reflectance and illumination, which could limit application in unusual lighting scenarios.

\section{Learning Colors with Existing Methods}
\label{sec:baselines_lc}

Here we dive into the challenges associated with synthesizing objects in desired colors using current state-of-the-art T2I diffusion or personalization methods. As elucidated in the paper, current T2I diffusion models offer users the ability to generate objects in desired colors by incorporating linguistic color descriptions in the prompts. Although these diffusion models have showcased remarkable capabilities in generating images from textual prompts, they encounter challenges in accurately reproducing specific colors. One of the primary reasons for this limitation is the broad spectrum of colors encompassed by linguistic color names (e.g., pink, blue, green), which can represent numerous combinations of hues and shades. For example, the color blue alone encompasses various shades such as navy blue, sky blue, royal blue, cobalt blue, and denim blue. Consequently, the generated images may not exactly match the intended color. As depicted in the examples presented in Fig.~\ref{fig:baselines_lc}(a) and Fig.~\ref{fig:baselines_lc}(c), even when prompted with standard color names, the stable diffusion model~\cite{Rombach_2022_CVPR_stablediffusion} and Rich-Text method~\cite{ge2023richtext} struggle to distinguish between different color variants.

\begin{figure*}[htp]
\centering
\includegraphics[width=0.9998\textwidth]{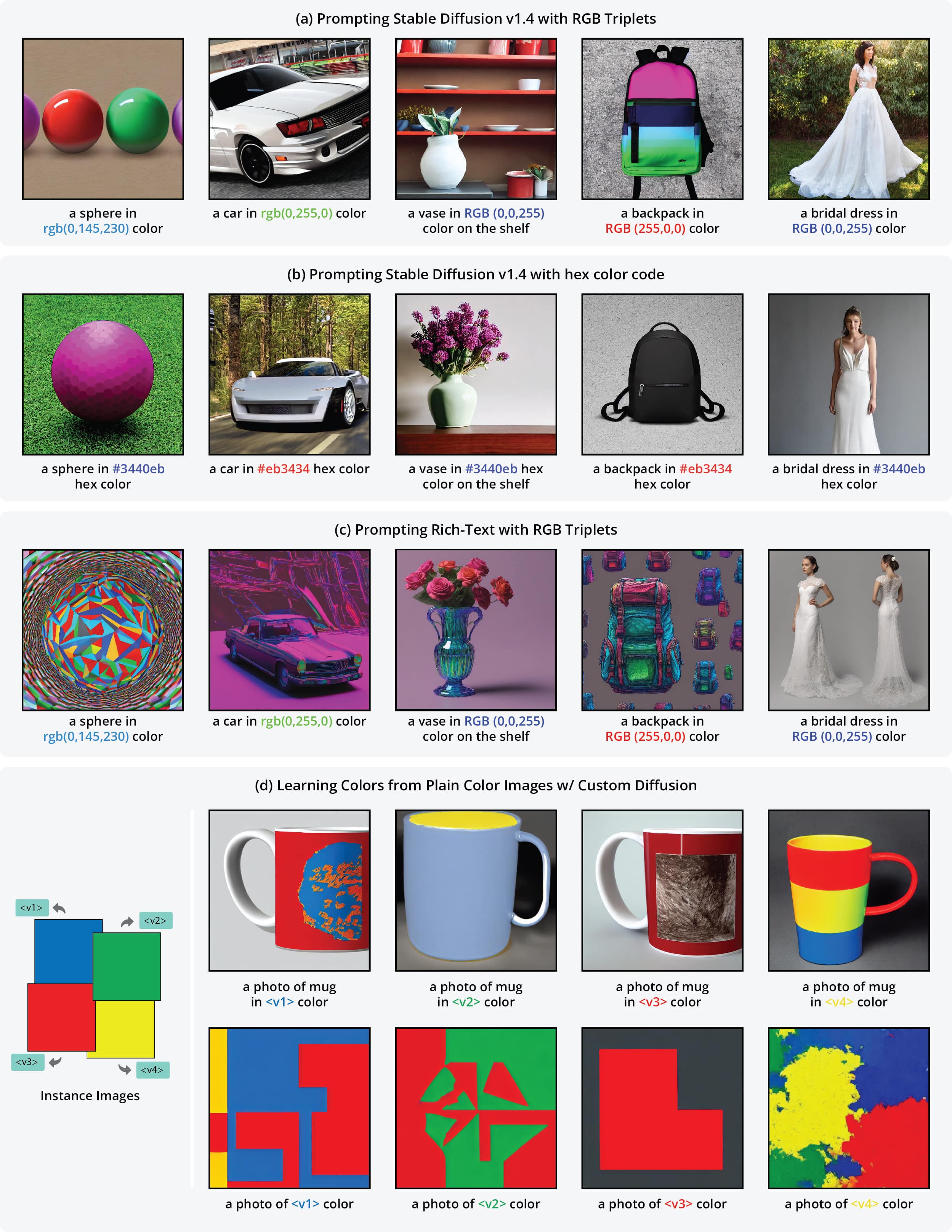}
\caption{
Visualizing failure compositions with existing methods. It can be noticed from the results that \textit{Stable Diffusion fails to comprehend \textbf{(a)} \textbf{(c)} RGB Triplets and \textbf{(b)} hex color code directly}. Whereas, multi-concept personalization method i.e., \textbf{(d)} Custom Diffusion struggles to learn color embeddings from plain images, which results in color intermixing in the generated outputs.
}
\label{fig:baselines_lc}
\end{figure*}

Another method of specifying precise colors is by including exact color values or hex color codes in the prompts to synthesize objects. The results depicted in Fig.~\ref{fig:baselines_lc}(b) reveal that the diffusion model struggles to interpret hex color codes or RGB values directly, as these models lack embeddings for such representations. Next, we consider T2I personalization methods to learn color embeddings based on new text tokens. We use well-established T2I personalization baselines, including TI~\cite{textual_inversion}, DB~\cite{ruiz2022dreambooth}, and CD~\cite{kumari2022customdiffusion}. First, we learn color embeddings from fully-colored images using RGB/hex values. However, as shown in Fig.~\ref{fig:baselines_lc}(d), these methods employ a naive transfer learning approach. While they are able to learn shapes or objects, they fail to effectively learn from plain color images. To gain further insight into this issue, we extract attention maps from the final timestep of the training process, as illustrated in Fig.~\ref{fig:supp_attn_maps}. It is evident from Fig.~\ref{fig:supp_attn_maps}(a) that the employed personalization methods struggle to focus on the color region when the entire image is colored. To address this limitation, we generate basic 2D/3D shapes using the RGB/hex values. Surprisingly, with this training setup, we observe that the employed methods, while still performing poorly, show some focus on the color regions. However, as demonstrated in the main paper, these methods tend to mix colors due to non-aligned token initialization.

\begin{figure*}[htp]
\centering
\includegraphics[width=0.9999\linewidth]{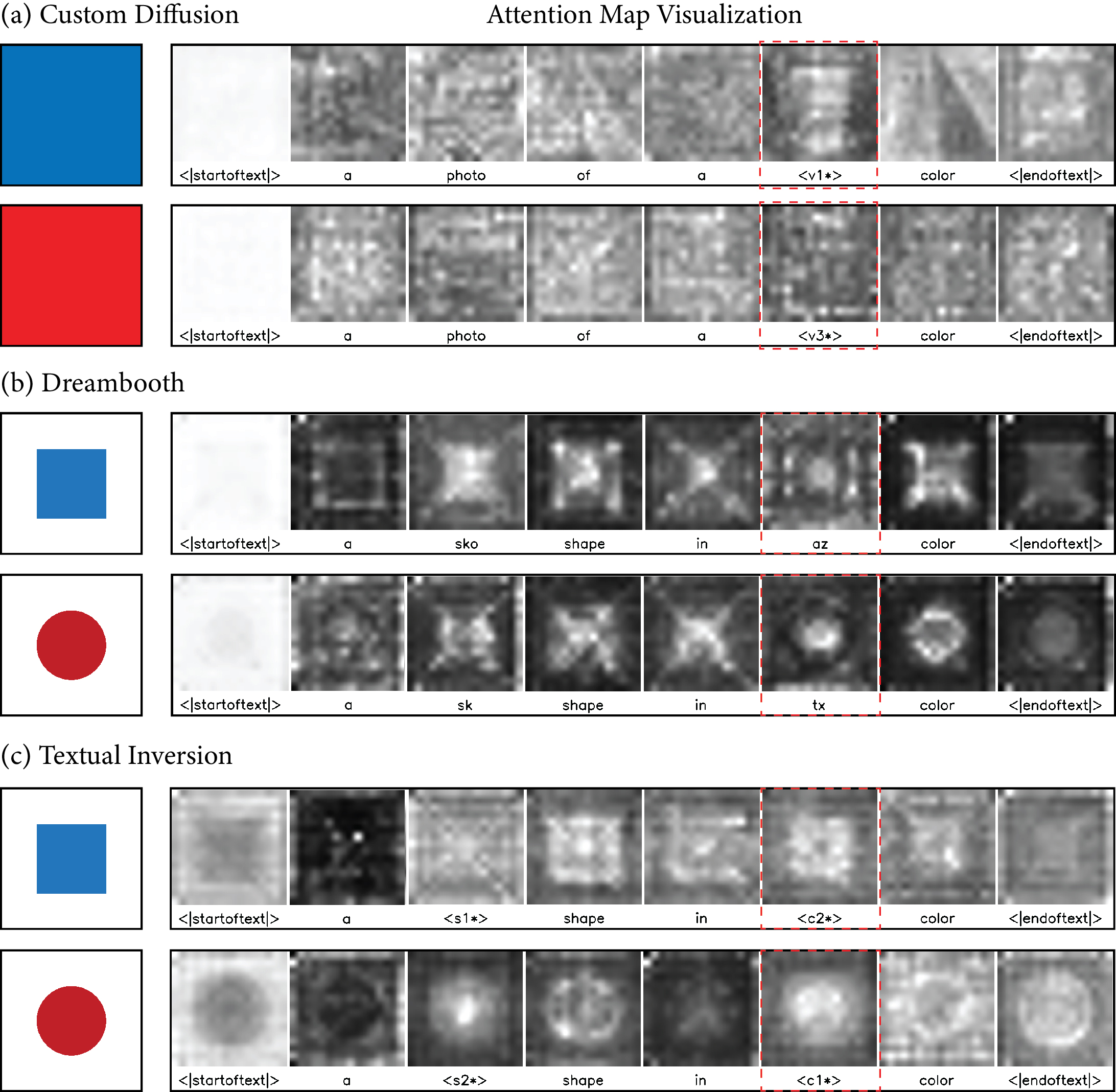}
\caption{Visualizing cross-attention maps from last time-step. \textbf{(a) Custom Diffusion}---fails to focus on the color in case of fully colored images. In particular, \textit{<v1*>} and \textit{<v3*>} which are supposed to learn colors, are not precisely learning. \textbf{(b) Dreambooth}---with basic colored shapes in the train images, the shape and color text tokens i.e., \textit{sko}, \textit{sk}, \textit{az}, and \textit{tx} are focusing on the color region, however, overlapping with the other tokens. Similarly, in \textbf{(c) Textual Inversion}, the new text-tokens \textit{<s1*>}, \textit{<s2*>}, \textit{<c1*>}, and \textit{<c2*>} are overlapping with other tokens. This can be one of the reasons which lead to the inaccurate color syntheses in the generated images.}
\label{fig:supp_attn_maps}
\end{figure*}

\section{Experiments}
\label{sec:expr_details}

\subsection{Implementation Details}
We train our \ourmethod with a batch size of 1 and a learning rate of $10^{-5}$. For coarse-color learning, we train the model for 1500 steps. However, we increase the training steps to 6000 steps for fine-grained color learning. It is important to mention that, we follow the same training schema for learning coarse and fine-grained colors as discussed in the section 3, for analyzing the disentanglement between the shapes and colors, along with the transferability of colors to unknown shapes/objects. To ensure faster convergence, we only back-propagate the valid regions' loss combined with the cross-attention alignment loss to improve learning.

\begin{figure*}[t]
\centering
\includegraphics[width=0.999\linewidth]{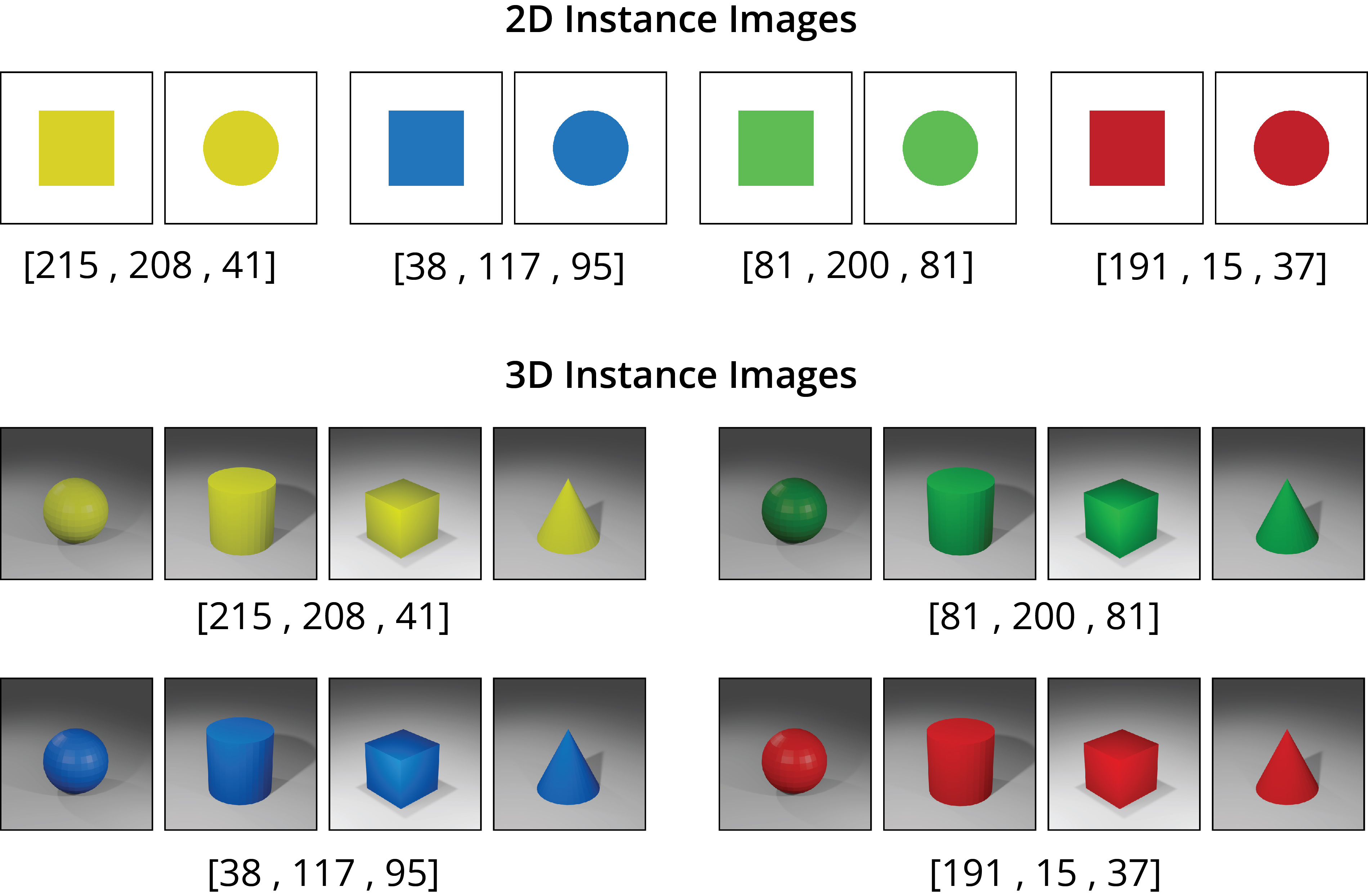}
\caption{2D and 3D instance images for the coarse-grained color learning task}
\label{fig:dataset_coarse}
\end{figure*}

\subsection{Dataset Details}
\label{sec:dataset_details}
As discussed in the previous section, colors have a countless range of combinations, due to different hues and shadings. Therefore, it is also crucial to prepare training-data/instance-images in the precise desired color. To handle this challenge, we introduce a data synthesizer in our \textit{\ourmethod}, which acts as a processing step in the pipeline. In particular, our method is capable of creating basic 2D and 3D shapes, given the desired RGB-Triplet, hex-Code, or Color-Coordinates. In 2D-Instances, rectangle and circle shapes are used, whereas, in the case of 3D, five simple 3D-shapes are used including \textit{sphere}, \textit{cube}, \textit{cylinder}, \textit{cone}, and \textit{hexagon}. It is important to mention here that 3d shapes provide more accurate representation, allowing for better understanding of color variations in different spatial dimensions. Therefore, we ensure that our \textit{\ourmethod} can learn color embeddings with both the 2D and 3D instance images. For 3D instance image creation, initially object files containing a scene graph with the shape positioned in the center of the plane, with three directional area lighting to ensure appropriate visibility, are created in the Blender. Our \textit{\ourmethod} can render these shapes, given the RGBs in real-time.

\begin{figure*}[htp]
\centering
\includegraphics[width=0.9999\linewidth]{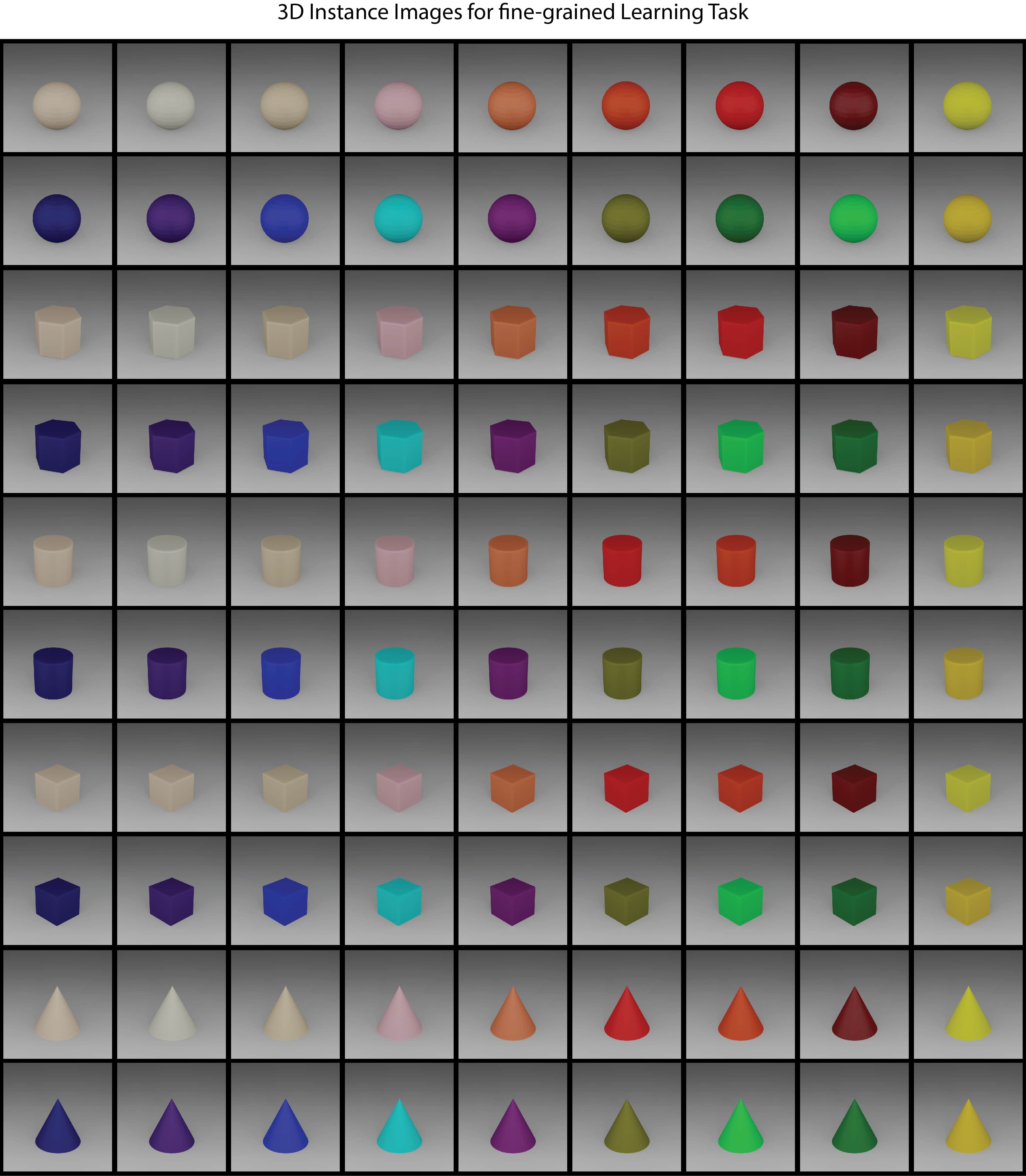}
\caption{3D instance images for the fine-grained color prompt learning task.}
\label{fig:dataset_fine}
\end{figure*}

For color prompt learning, we design two color learning tasks: (i) coarse-grained color learning, which contains four basic colors---red, green, blue, and yellow, and (ii) fine-grained color learning which covers $18$ colors related to less common color names, including 'salmon','beige', etc. The sample instance 2D and 3D images for coarse-grained color learning task are shown in the Fig.~\ref{fig:dataset_coarse}, and the 3d instance images are demonstrated in Fig.~\ref{fig:dataset_fine} and the list of the color names along with corresponding RGB values are enlisted in Table~\ref{tab:rgb_codes}.

\begin{table}[t]
  \centering
  \caption{RGB values of colors used to generate coarse and fine-grained color sets. Coarse-grained color learning tasks covers four basic colors, whereas, fine-grained color learning task includes 18 colors.}
  \resizebox{0.3\columnwidth}{!}{%
  \begin{tabular}{@{}ll@{}}
    \toprule
    Color & RGB Code \\
    \midrule
    \multicolumn{2}{c}{Coarse-grained Color Set} \\
    \hline
    Red & 191, 15, 37 \\
    Green & 81, 200, 81 \\
    Blue & 38, 117, 195\\
    Yellow & 215, 208, 41\\
    \hline
    \multicolumn{2}{c}{Fine-grained Color Set} \\
    \hline
    Red & 255, 0, 0 \\
    Maroon & 128, 0, 0 \\
    Orange  & 255, 165, 0 \\
    Coral & 255, 127, 80 \\
    Pink & 255, 192, 203 \\
    Green & 0, 128, 0 \\
    Lime & 0, 255, 0 \\
    Olive & 128, 128, 0 \\
    Blue & 0, 0, 255\\
    Navy & 0, 0, 128 \\
    Cyan & 0, 255, 255 \\
    Turquise & 64, 255, 208 \\
    Indigo & 75, 0, 130 \\
    Purple & 128, 0, 128 \\
    Yellow & 255, 255, 0\\
    Gold & 255, 215, 0 \\
    Bisque & 255, 228, 196 \\
    Wheat & 245, 222, 179 \\
    Beige & 245, 245, 220 \\
    \bottomrule
  \end{tabular}
  }
  \label{tab:rgb_codes}
\end{table}

\begin{figure*}[htp]
\centering
\includegraphics[width=0.9999\textwidth]{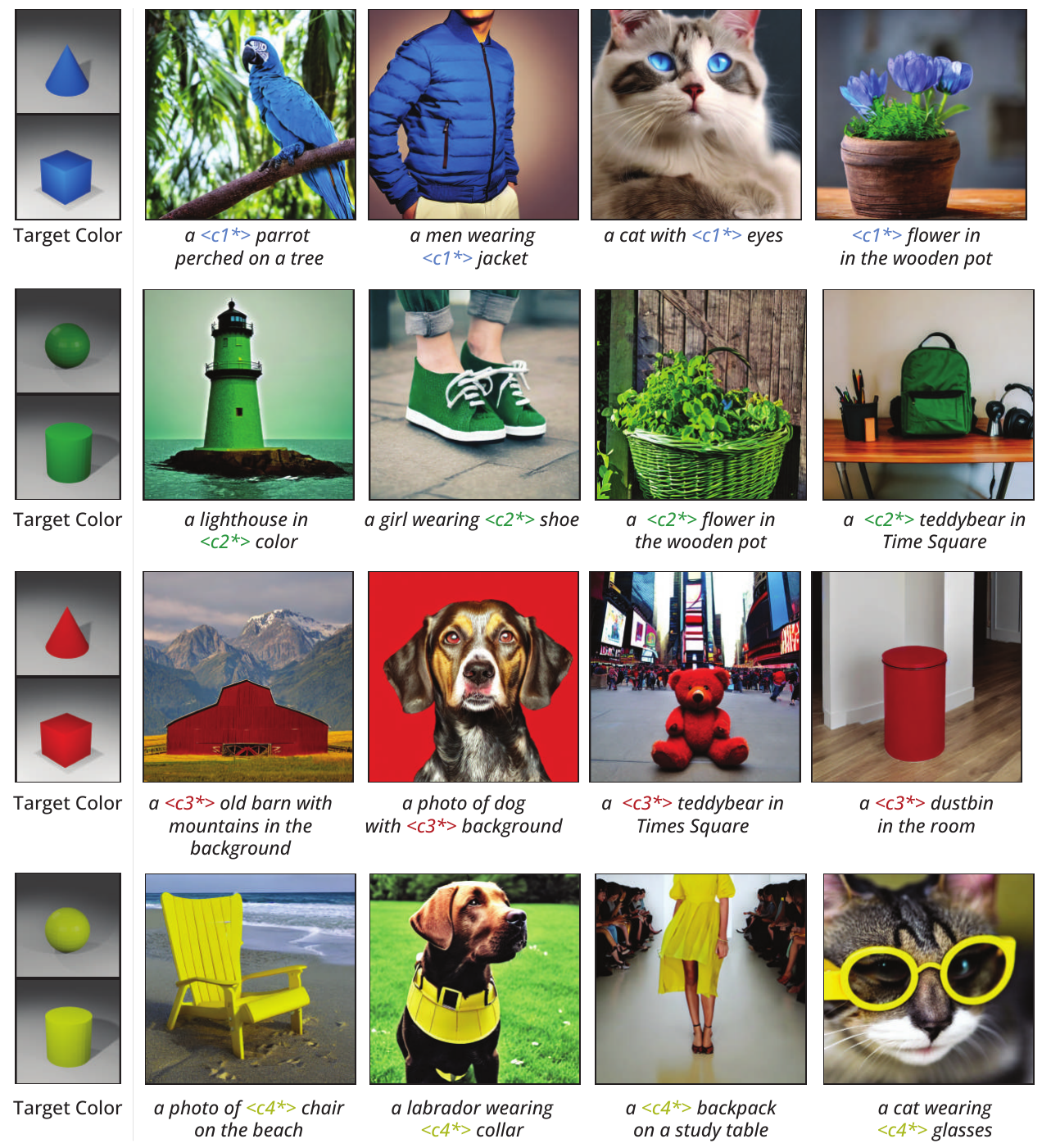}
\caption{Qualitative color generation results of the coarse-grained color learning task.}
\label{fig:coarse_results}
\end{figure*}

\subsection{Additional Qualitative Results}
The results our method \textit{\ourmethod} for the coarse color and fine color learning tasks are demonstrated in Fig.~\ref{fig:coarse_results}, and Fig.~\ref{fig:fine_results}, respectively. In addition to generating high-quality images, \textit{\ourmethod} also showcases effective personalization of diverse elements, including customizing attire like clothing, footwear, gloves, and glasses, as well as toys and objects within different settings.
In the next step, we carefully designed the prompts to evaluate the color transferability in terms of consistency and fidelity to a wide range of attributes. These attributes span from image background color customization to personalized elements such as clothing, eye, hair colors, as well as other objects like chair, dustbin, etc.

\begin{figure*}[t]
\centering
\includegraphics[width=0.999\linewidth]{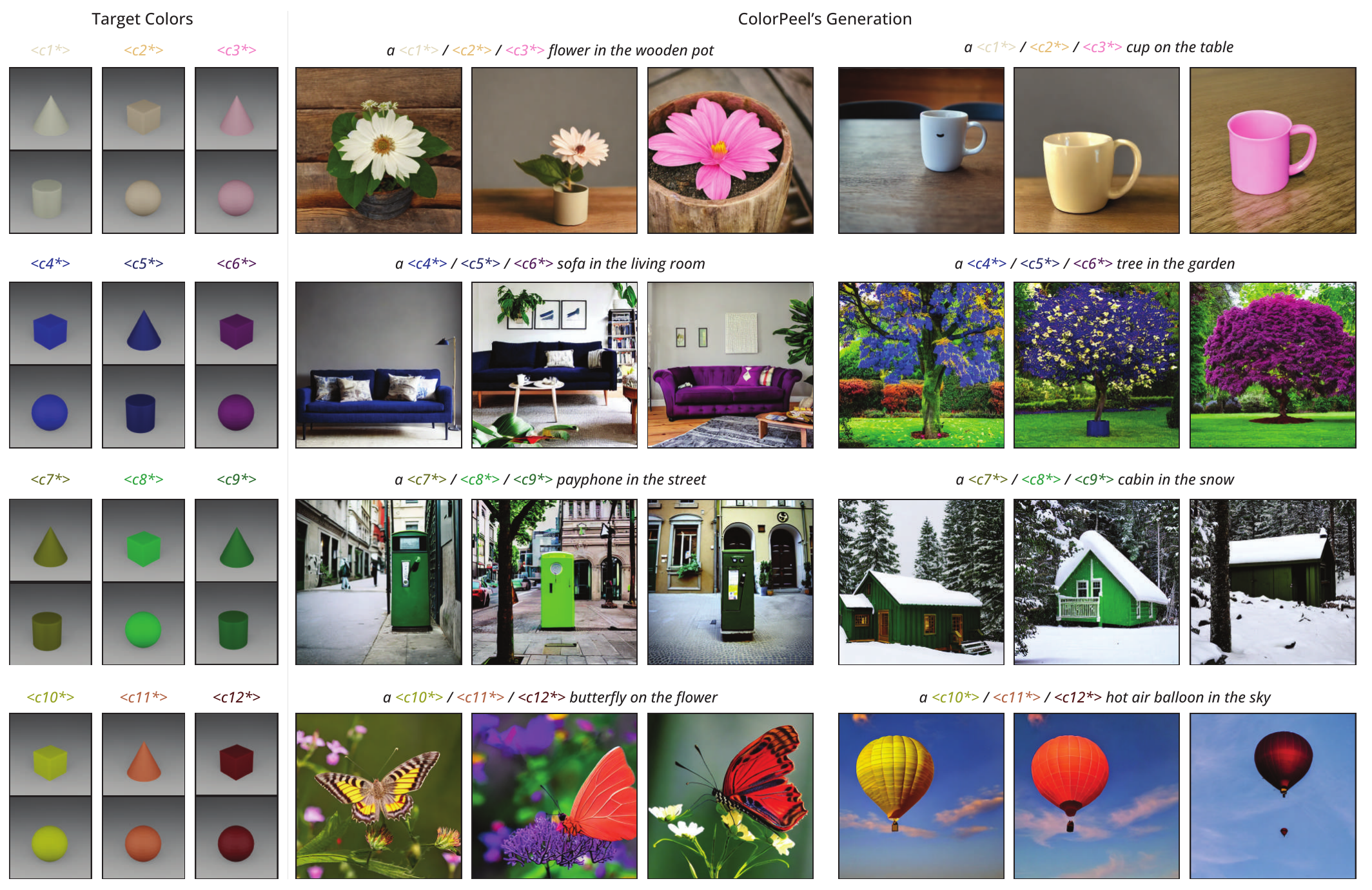}
\caption{Qualitative color generation results of the fine-grained color learning task.}
\label{fig:fine_results}
\end{figure*}

\subsection{Prompt Templates}
\label{subsec:prompt_templates}
\minisection{Evaluation prompt templates.}
Here is the list of text prompts used in evaluating our proposed method and comparing it with the baseline methods i.e., Stable Diffusion, Custom Diffusion~\cite{kumari2022customdiffusion}, Textual Inversion~\cite{textual_inversion}, Rich Text~\cite{ge2023richtext}, and Dreambooth~\cite{ruiz2022dreambooth}.

\begin{itemize}
    \item a \{color\} bowl on the table
    \item a \{color\} bowling ball in a bowling alley
    \item a \{color\} plate on the table
    \item a \{color\} vase on the shelf
    \item a women wearing \{color\} pants
    \item a \{color\} teddy-bear in Time Square
    \item a \{color\} snooker ball on the table
    \item a \{color\} parrot perched on a tree
    \item a \{color\} sofa in living room
    \item a \{color\} rose blooming in a wooden pot

\end{itemize}

\minisection{Training prompt templates.}
Initially, we tried optimizing the new text-tokens with multiple training prompt examples, enlisted below.

\begin{itemize}
    \item a photo of <s*> shape in <c*> color 
    \item a <s*> shape in <c*> color
    \item a <c*> colored <s*> shape
    \item a photo of <c*><s*>
\end{itemize}

 \noindent However, \textit{\ourmethod} achieved better results with single prompt i.e., "\textit{a photo of <c*> shape in <s*> shape}" for each instance image, where <s*> and <c*> are new text-tokens, corresponding to shape and color, respectively.

\begin{figure*}[t]
\centering
\includegraphics[width=0.9999\textwidth]{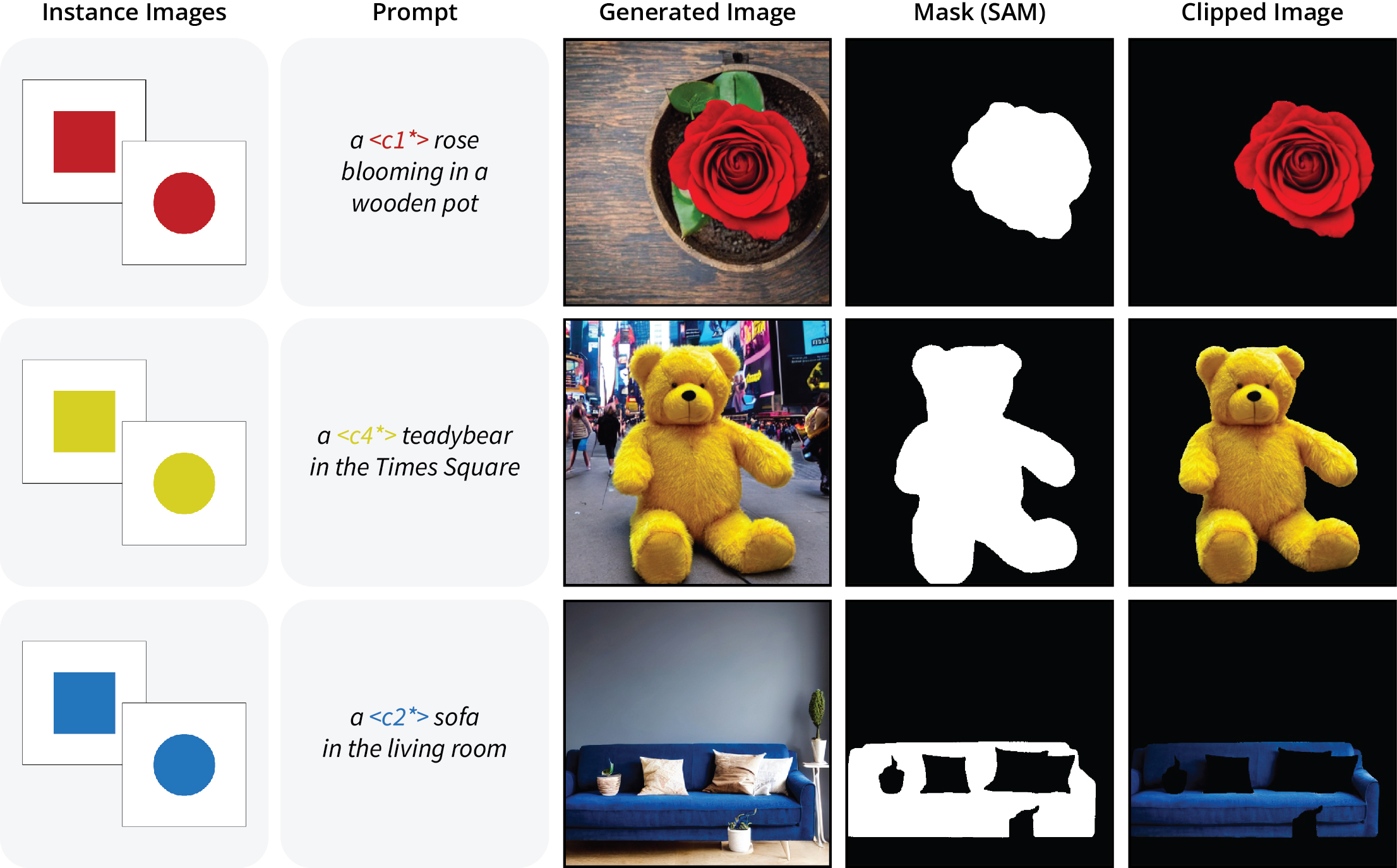}
\caption{Mask Generation for quantitative evaluation. Given an image generated with a prompt, we compute a mask using Segment Anything model~\cite{kirillov2023segment_anything_sam} in order to consider only those pixels that belong to the object.}
\label{fig:gen_mask}
\end{figure*}

\begin{figure}[t]
\centering
\includegraphics[width=0.9999\textwidth]{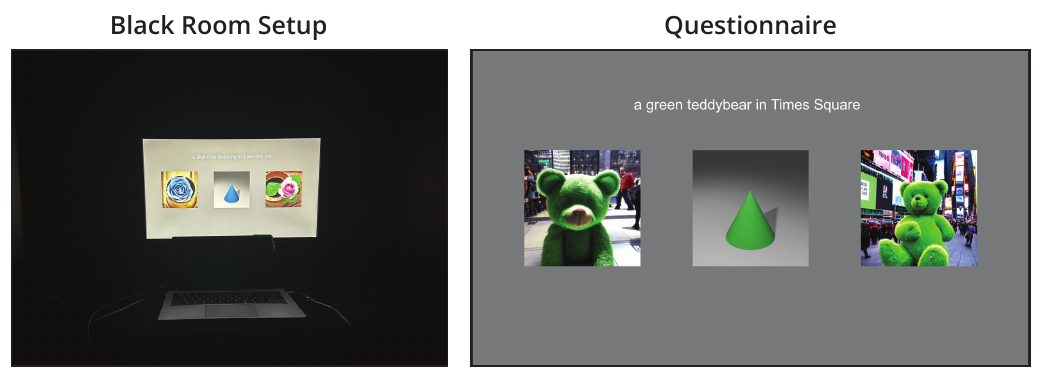}
\caption{Setup of our Human Study. \textit{Left}: Observers were sitting in a black room, where the only light was the one provided by the monitor. \textit{Right}: The background of the monitor screen was fixed to middle gray, and the observers needed to select which image ---left or right--- was a better match considering the prompt and the color shown in the central image.
}
\label{fig:hum_exp}
\end{figure}

\subsection{Evaluation pipelines}
\label{subsec:eval}
Fig.~\ref{fig:gen_mask} shows the masking computation. An image is generated given a prompt. Then, the Segment Anything Model \cite{kirillov2023segment_anything_sam} is used to compute the mask of the generated object ---in this figure the red rose, the yellow teddy-bear, or the blue sofa---. This mask is then used to consider only the object's pixels for the computation of the metrics. This said, for some objects they might be some small parts that are not supposed to have the desired color; for example the eyes of the teddy-bear. For this reason, in the main paper we computed some measures considering also the 10\%, or 50\% of most correct pixels inside the mask. Let us remind the reader that our method outperformed all the others in all metrics at under all conditions.

\begin{figure*}[hbtp]
\centering
\includegraphics[width=0.98\linewidth]{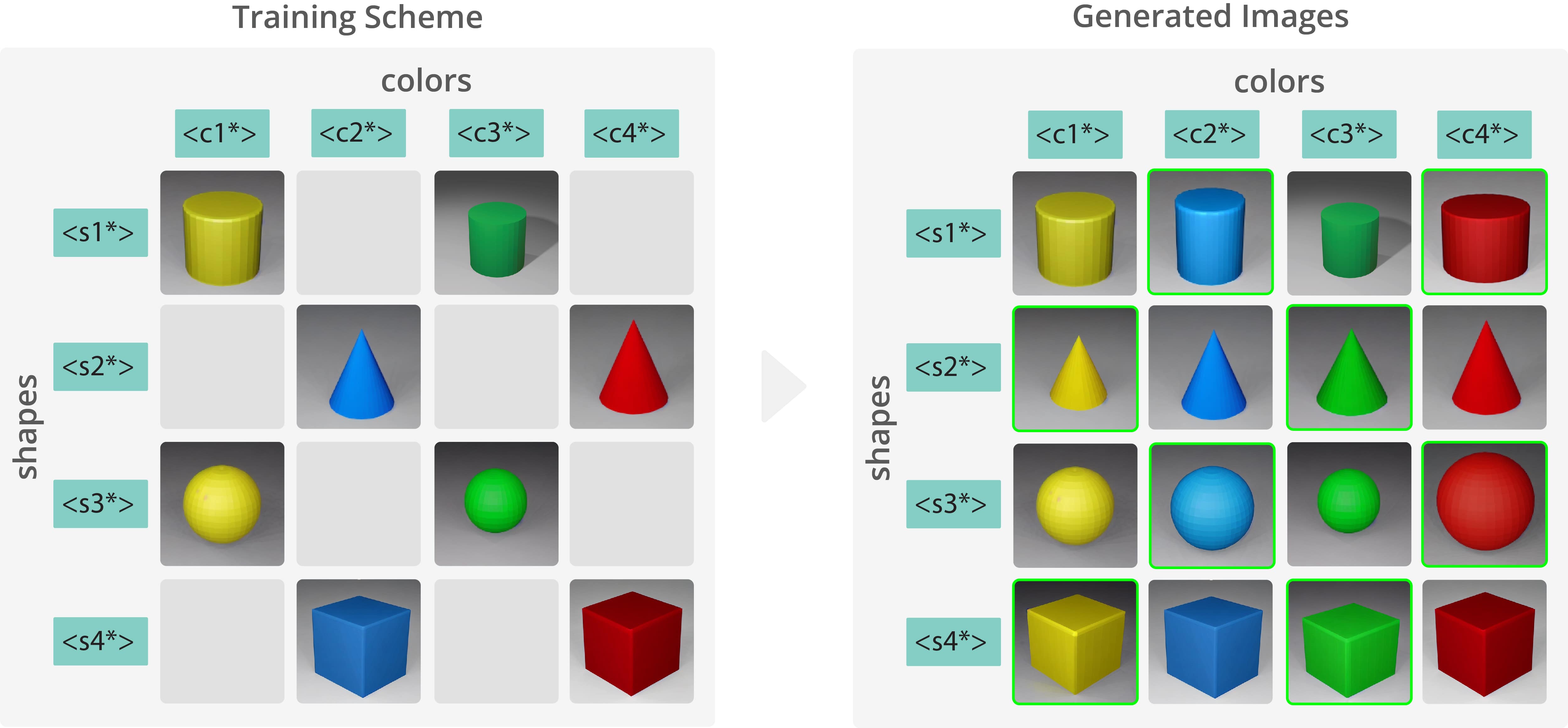}
\caption{Training scheme and generating unknown shapes. The instance prompt---"\textit{a photo of <s*> shape in <c*> color}" is used to reproduce the shapes with unknown colors. The generated images are outlined in green border.}
\label{fig:4x4_training_scheme}
\end{figure*}

\subsection{User study}

Fig.~\ref{fig:hum_exp} shows our user's study setup. In the left, we see a picture of the room, that was completely black and the monitor set to sRGB. The monitor ---Fujitsu B-24-8--- was the only light source during the experiment. Observers were seated approximately 60 cm away from the monitor to ensure a 7-degree visual angle.
The monitor background was set to middle gray. The monitor displayed a prompt and a central image with the reference color. Left and right from the central image we randomly showed the results for our method and a competing one. The observer needed to select which image from the two represented better the prompt given the color in the middle image. An example of the set-up as seen by the observer is shown in the right part of Fig.~\ref{fig:hum_exp}. A total of $15$ observers participated in the study, and none of the authors took part in the study.

\subsection{Verification of the Color Prompt Learning}
We conduct experiments over coarse-grained concepts using the training schema as shown in Fig.~\ref{fig:4x4_training_scheme} to analyze the learning of color prompts from given colored shapes and their transferability to other objects. To evaluate if our method \textit{\ourmethod} is correctly disentangling the colors from shapes and can transfer to the unknown ones, as devised in training schema, we learn colors and shapes given in coarse-grained training set in $\colortokeni$ and $\shapetokeni$ text-tokens, respectively. The results are illustrated in Fig.~\ref{fig:4x4_training_scheme}, we showcase that with only eight images in the 4$\times$4 training scheme, \ourmethod can successfully infer the geometries not included in the training set. That further proves the effectiveness of our method \ourmethod.

\subsection{Image Editing}
Here we demonstrate a few examples of image editing following the P2P method~\cite{hertz2022prompt} with our \textit{\ourmethod}. The corresponding image editing results are shown in Fig.~\ref{fig:image_editing}, where we successfully modified the color of the objects to our learned colors.

\begin{figure*}[hbtp]
\centering
\includegraphics[width=0.97\linewidth]{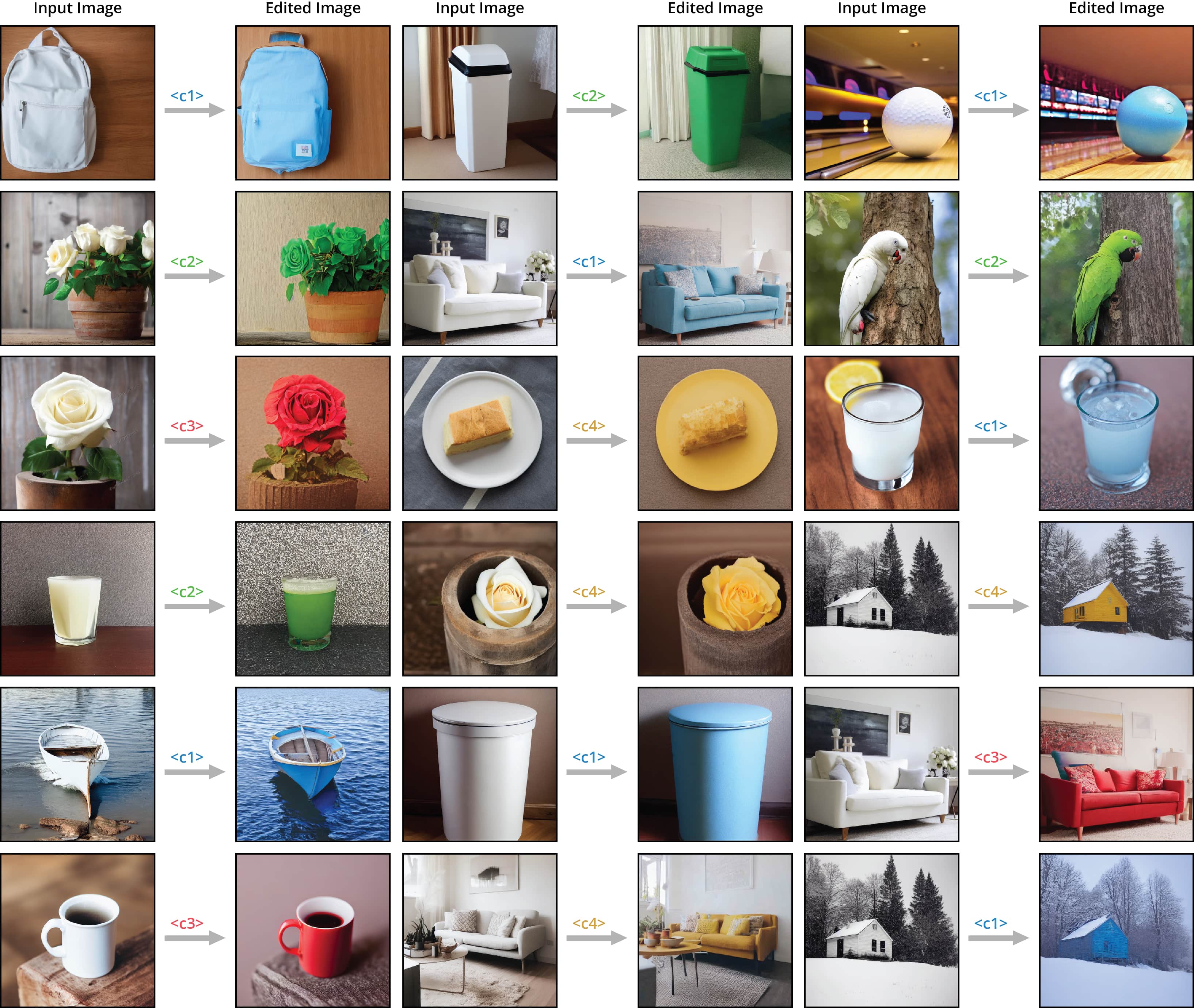}
\caption{Demonstrating the results of image editing. <c1*>, <c2*>, <c3*>, and <c4*> corresponds to the embeddings of newly learned colors for the coarse-grained color learning task. The results show that the object colors can be modified given our learned embeddings.}
\label{fig:image_editing}
\end{figure*}

\subsection{Color Interpolation}
\label{subsec:color_interpolation}
Here we demonstrate a few examples of linear interpolation between two newly learned color tokens. The results are shown in Fig.~\ref{fig:color_interpolation} which shows that our \textit{\ourmethod} can represent the colors continuously between the learned color prompts, which ultimately can avoid the training for new colors.

\begin{figure*}[hbtp]
\centering
\includegraphics[width=\linewidth]{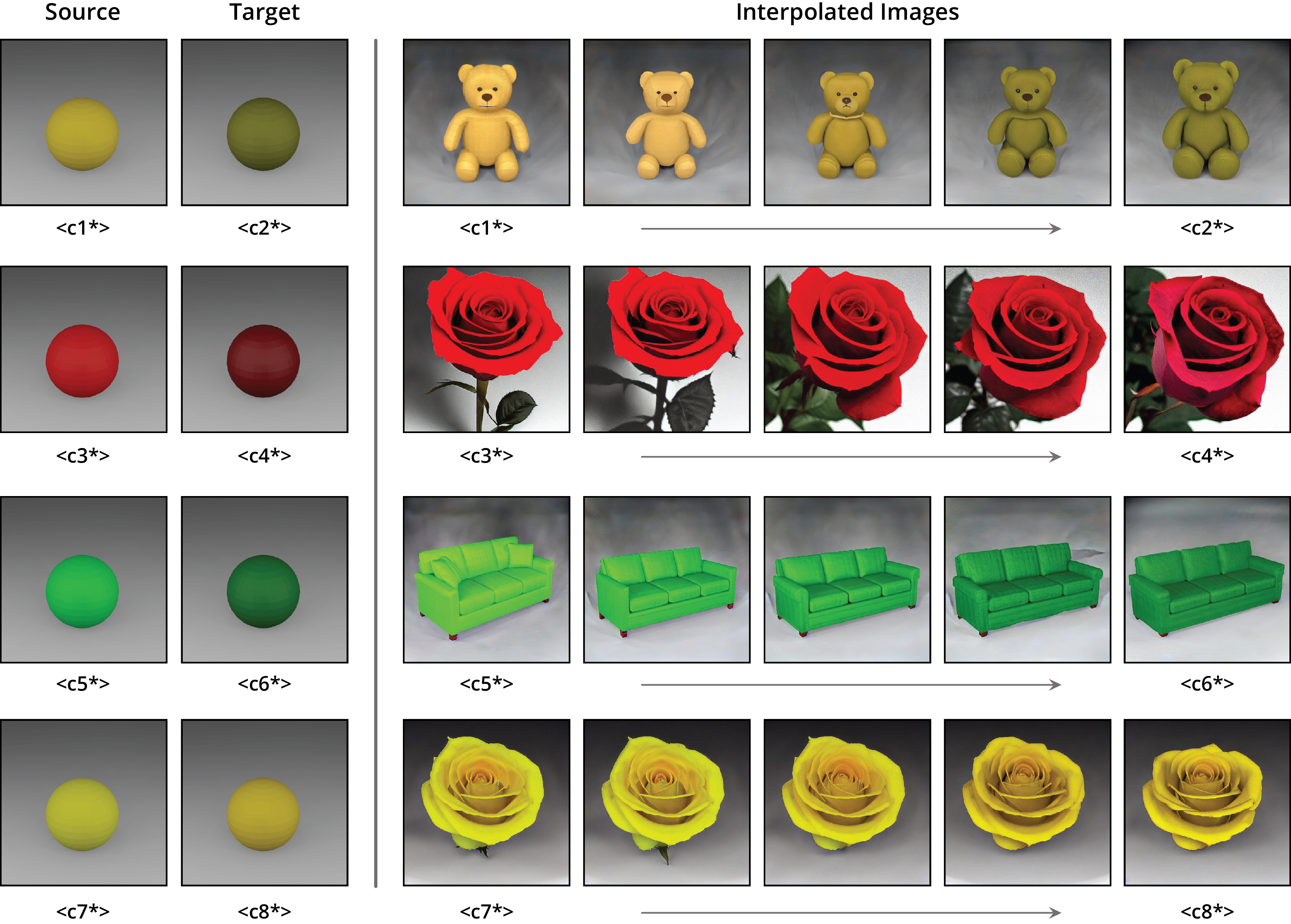}
\caption{Linear interpolation between two color tokens.}
\label{fig:color_interpolation}
\end{figure*}

\subsection{Additional Qualitative Comparison}
\label{subsec:additional_comp_supp}
We show the additional qualitative comparison between the generated images from baselines including Stable Diffusion (SD), Rich-Text, Textual Inversion (TI), Dreambooth (DB), and Custom Diffusion (CD) against our method \ourmethod. The results are shown in Fig.~\ref{fig:qual_comparison} which shows that our \textit{\ourmethod} can learn and transfer better photo-realistic colors as compared to the baselines.

\begin{figure*}[hbtp]
\centering
\includegraphics[width=\linewidth]{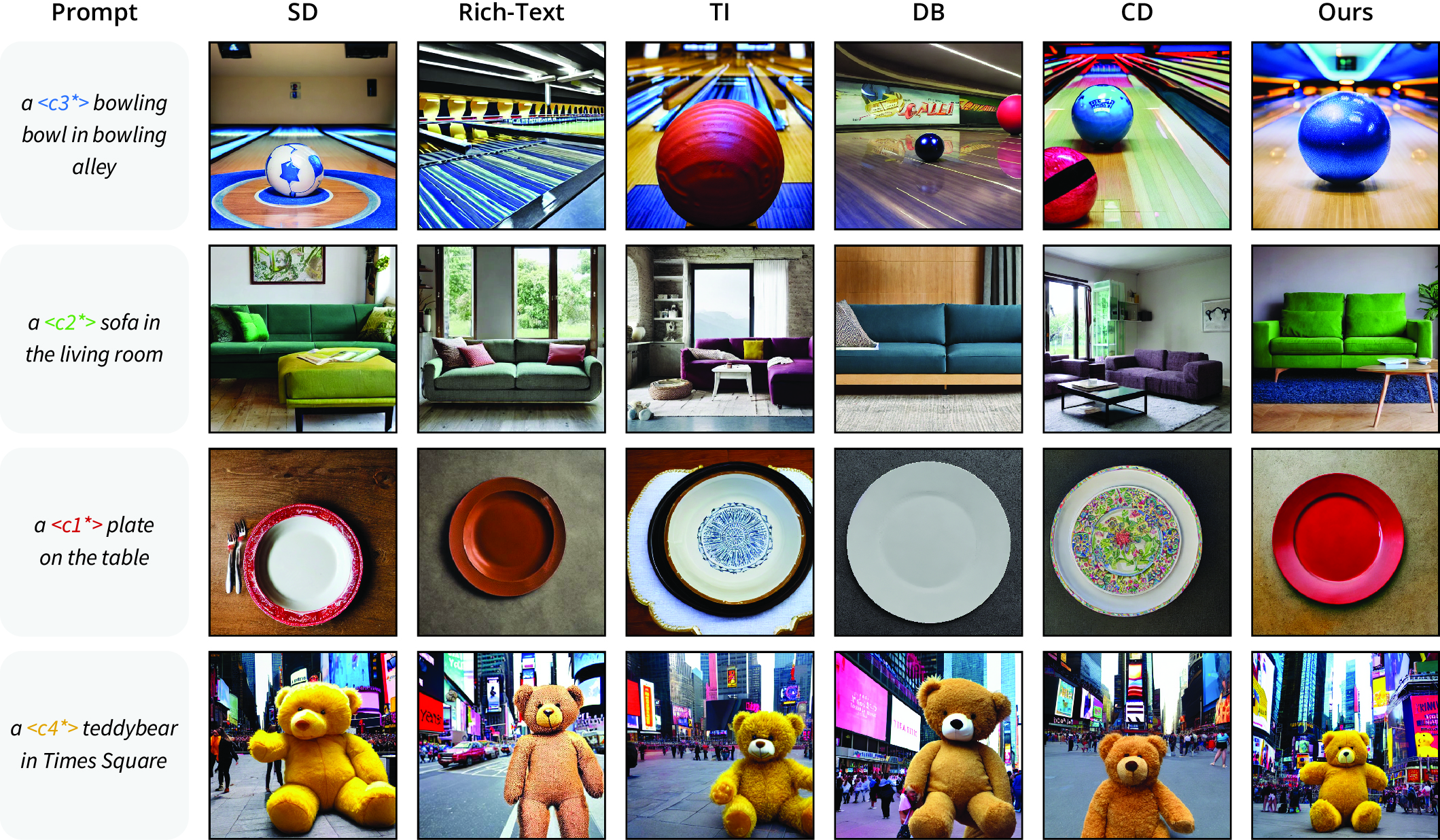}
\caption{Demonstrating qualitative comparison of the generated images with the baselines and our method.}
\label{fig:qual_comparison}
\end{figure*}

\begin{figure*}[hbtp]
  \centering
  \includegraphics[width=0.90\linewidth]{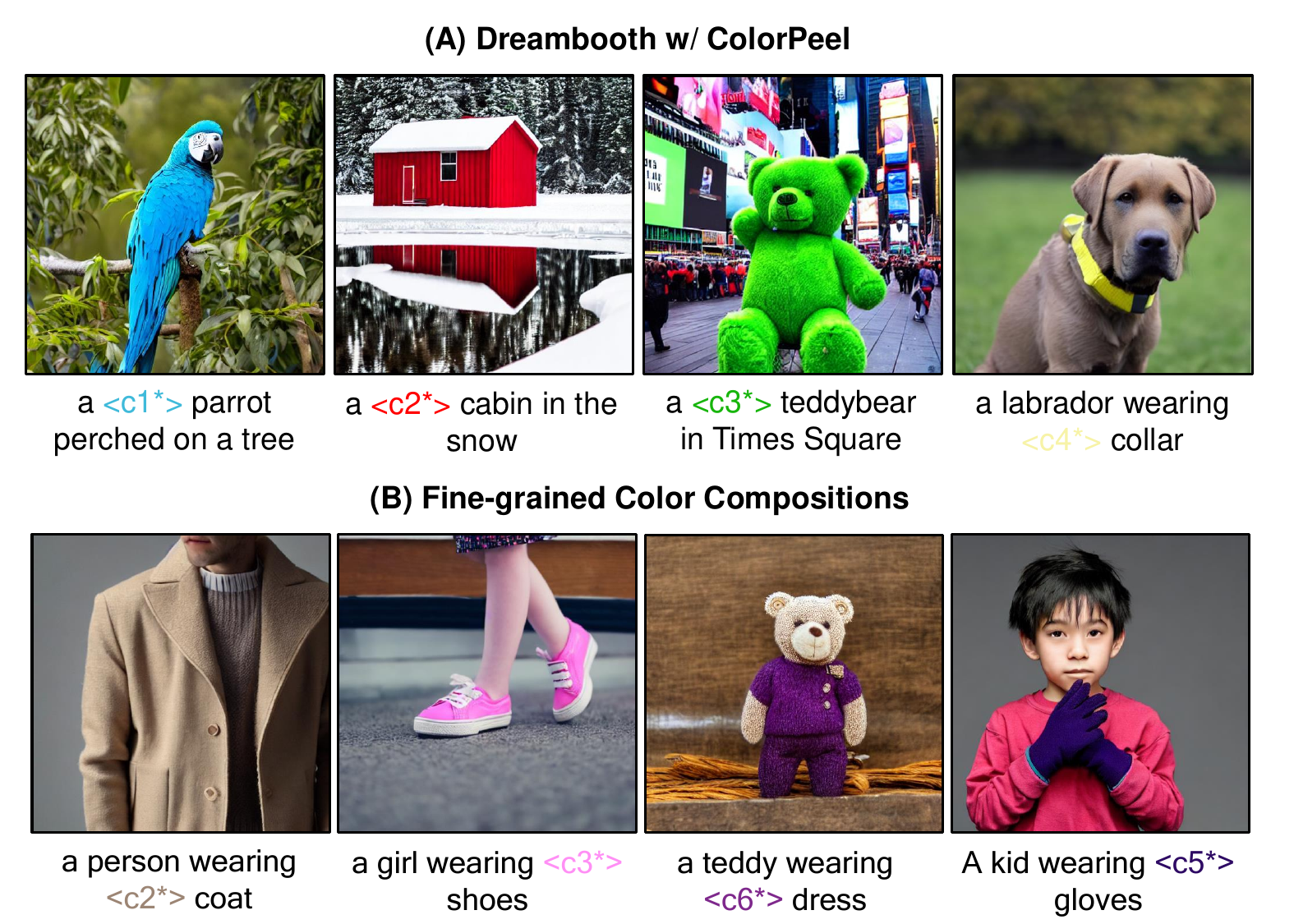}
  \caption{Demonstrating results of (A)  DB trained using \ourmethod and (B) additional fine-grained color generation.}
  \label{fig:fine_grained_gen_}
\end{figure*}

\section{Ablation Study}
\label{sec:abl_study}
We ablate various components of our method to show its contribution, which are demonstrated in Fig.~\ref{fig:ablations_expanded}. Firstly, we remove cross attention loss (CAA) and train the model with the default baseline settings. The results (see Fig.~\ref{fig:ablations_expanded}a) demonstrate that the model fails to disentangle the color from the shape, and replicates the shapes while ignoring the target prompt. We also notice that the default transformation techniques (resize, zoom, etc.) in the baseline are highly influencing the reconstruction performance, resulting in inaccurate generation. Secondly, we use fully weighted cross attention loss combined with the reconstruction loss to train the model. The results (see Fig.~\ref{fig:ablations_expanded}b) demonstrate that model ensures efficient learning and transferability of colors, however, model fails in accurate shape reconstruction. Thirdly, we trained the model by scaling down the lambda to 0.7 in CAA loss which improves the color transferability and shape reconstruction as compared to fully weighted CAA loss. Lastly, we show the results of our \ourmethod where we train the model by setting lambda to 0.2 in CAA loss. With this setting, our method achieves comparatively better performance in terms of color fidelity and consistency, and shape reconstruction.

\begin{figure*}[t]
\centering
\includegraphics[width=0.9999\textwidth]{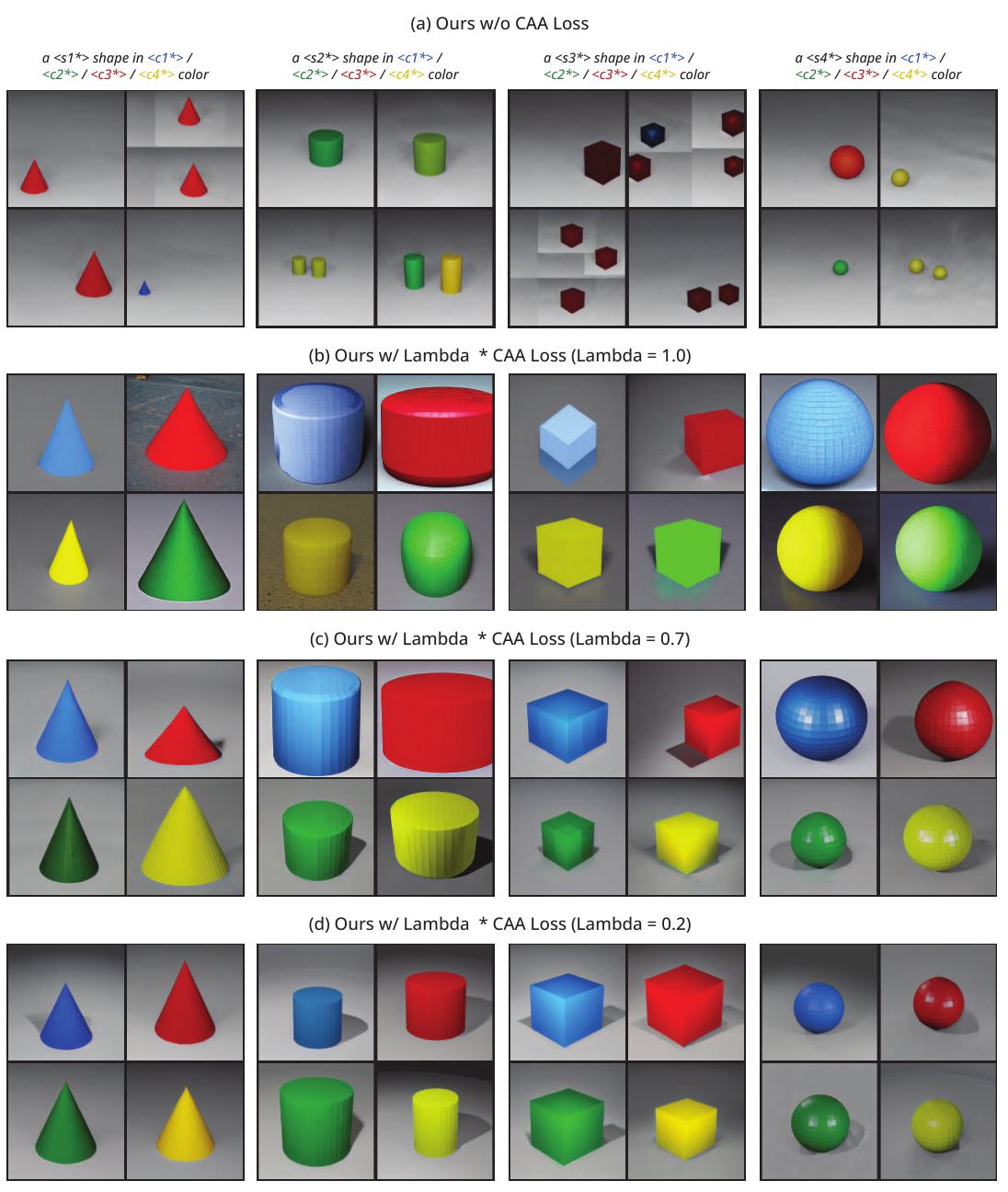}
\caption{\textbf{Ablation Study.} We ablate various components of our method \ourmethod to demonstrate their contributions.}
\label{fig:ablations_expanded}
\end{figure*}

\end{document}